\documentclass[a4paper,11pt,titlepage,oneside]{article}
\pdfoutput=1
\usepackage[utf8]{inputenc}
\usepackage[T1]{fontenc}
\usepackage{microtype}
\usepackage[english]{babel}
\usepackage[dvipsnames]{xcolor}
\usepackage{graphicx}
\usepackage{hyperref}
\hypersetup{colorlinks=true,allcolors=black}
\usepackage[all]{hypcap}        

\usepackage{amsmath}
\usepackage{amsfonts}
\usepackage{amssymb}
\usepackage{amsthm}
\usepackage{mathtools}
\usepackage{thmtools}

\usepackage[tighter]{newtxtext}
\usepackage[varg,subscriptcorrection]{newtxmath}

\usepackage{booktabs}
\usepackage{longtable}
\usepackage{tabularx}

\usepackage{notestikz}
\usepackage{tikzpagenodes}      

\usepackage[frozencache=true,cachedir=.]{minted}
\setminted{fontsize=\normalsize,baselinestretch=1}
\definecolor{bg}{rgb}{0.95,0.95,0.95}
\newmintedfile{turtle}{}
\newminted{turtle}{}
\newmintinline[ttl]{turtle}{}
\newmintinline[code]{text}{}
\newmintinline[scode]{text}{fontsize=\small}

\usepackage{notessymb}
\newcommand*{\C}[1]{\mathcal{#1}}
\newcommand*{\T}[1]{\mathtt{#1}}
\newcommand*{\TT}[1]{\operatorname{\T{#1}}}
\newcommand*\TC[1]{\text{\code{#1}}}
\newcommand*\TSC[1]{\text{\scode{#1}}}
\renewcommand*{\S}[1]{\mathsf{#1}}
\renewcommand*{\SS}[1]{\operatorname{\S{#1}}}

\newcommand*{\I}{\C{I}}
\newcommand*{\J}{\C{J}}

\newcommand*\ssC{\scriptscriptstyle{𝐂}}
\newcommand*\ssR{\scriptscriptstyle{𝐑}}

\newcommand*{\KB}{\C{KB}}
\newcommand*{\SROIQ}{\ensuremath{\C{SROIQ}}}

\begin{document}
\title{An Introduction to Symbolic Artificial Intelligence Applied to
  Multimedia}
\author{%
  Guilherme Lima,
  Rodrigo Costa,
  Marcio Ferreira Moreno\\[1\jot]
  {\small
    \url{Guilherme.Lima@ibm.com},
    \url{Rodrigo.Costa@ibm.com},
    \url{mmoreno@br.ibm.com}}\\[3\jot]
  IBM Research, Brazil}
\date{2019-11-08}

\maketitle
\begin{abstract}
  In this chapter, we give an introduction to symbolic artificial
  intelligence (AI) and discuss its relation and application to multimedia.
  We begin by defining what symbolic AI is, what distinguishes it from
  non-symbolic approaches, such as machine learning, and how it can used in
  the construction of advanced multimedia applications.  We then introduce
  description logic (DL) and use it to discuss symbolic representation and
  reasoning.  DL is the logical underpinning of OWL, the most successful
  family of ontology languages.  After discussing DL, we present OWL and
  related Semantic Web technologies, such as RDF and SPARQL.  We conclude
  the chapter by discussing a hybrid model for multimedia representation,
  called Hyperknowledge.  Throughout the text, we make references to
  technologies and extensions specifically designed to solve the kinds of
  problems that arise in multimedia representation.
\end{abstract}

\tableofcontents
\cleardoublepage

\section{Introduction}
\label{sec:intro}

A classic problem in multimedia representation and understanding is the
\emph{semantic gap} problem.  It states that there is a big representational
gap between the audiovisual signals that compose multimedia objects and the
concepts represented by these signals.  For instance, the dominant color and
movement trajectory of a given set of pixels in a video clip, which are
low-level characteristics of the clip, do not provide much information about
the \emph{meaning} of the set of pixels---at least not to computers.  But
recent developments in artificial intelligence (AI) are changing that.

Backed by large training datasets, current machine learning methods are able
to extrapolate complex patterns from low-level multimedia data.  These
patterns are embodied in trained models which can be used to classify or
identify persons and objects with reasonable speed and accuracy in images,
audio clips, and video clips.  But being able to identify persons and
objects in multimedia data only solves half of the problem.  To emulate
human cognition and truly understand a scene---for instance, to determine
who is doing what and the consequences of these actions---computers need
additional information: they need common sense knowledge and domain
knowledge, and also the capacity to infer new knowledge from preexisting
knowledge.  This is where symbolic AI comes in.

\begin{figure}[h]
  \centering
  \begin{tikzpicture}[node distance=1em,
    w/.style={draw,white,thick,inner sep=0,outer sep=0},
    l/.style={white,font=\sffamily\bfseries\scriptsize}]
    \node[draw,anchor=south west,inner sep=0pt](marat) at (0,0)
    {\includegraphics[width=.8\textwidth]{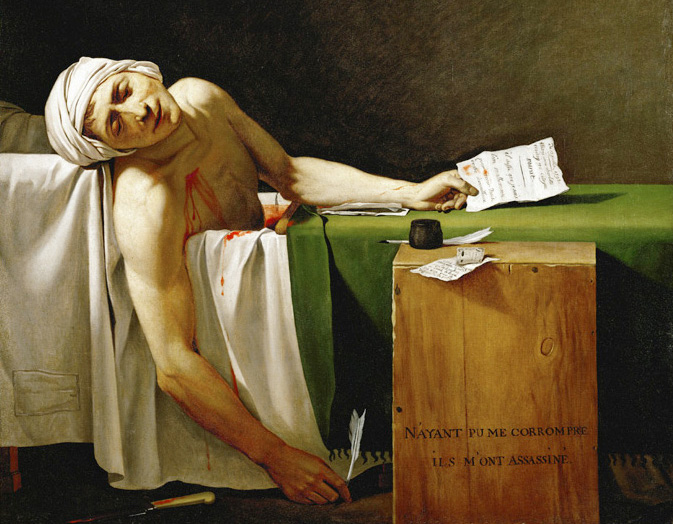}};
    \begin{scope}[x={(marat.south east)},y={(marat.north west)}]
      \clip(0,0) rectangle (1,1);
      \node[w,shape=ellipse,minimum width=6em,
      minimum height=1.25em,rotate=10](knife)at (.2,.025){};
      \node[l](knifelbl) at (.08,.12){Knife};
      \draw[white](knifelbl) to (knife);
      \node[w,shape=ellipse,minimum width=4em,minimum height=4em](head)
      at (.195,.79){};
      \node[l,above right=of head,yshift=-.5em](headlbl){Jean-Paul Marat};
      \draw[white](headlbl) to (head);
      \node[w,shape=ellipse,minimum width=1em,minimum height=1em](wound)
      at (.285,.66){};
      \node[l,above right=of wound,xshift=1em,yshift=1em]
      (woundlbl){Wound};
      \draw[white](woundlbl) to (wound);
      \node[w,shape=ellipse,minimum width=4em,minimum height=2.5em](letter)
      at (.75,.66){};
      \node[l,text width=8em,align=center,above=of letter](letterlbl)
      {Letter from\\Charlotte Corday};
      \draw[white](letterlbl) to (letter);
    \end{scope}
  \end{tikzpicture}
  \caption{\emph{The Death of Marat} (detail), by Jacques-Louis
    David, 1793.  (WikiMedia)}
  \label{fig:marat}
\end{figure}

The basic idea of symbolic AI is to describe the world, its entities, and
their relationships using a formal language---a language that can be
conveniently manipulated by computers---and to develop efficient algorithms
to query and deduce things from these formal descriptions.  To illustrate
the kind of applications enabled by the combination of symbolic AI and
multimedia consider Figure~\ref{fig:marat}.

Suppose we are given this picture and suppose the only thing we know about
it is what we can infer from the image.  We can see it depicts Jean-Paul
Marat (assuming we can identify him), a stab wound, a blood-tainted knife,
and a letter addressed to him and signed by Charlotte Corday (assuming we
can read the contents of the letter).  The analogy here is that we have
extracted these information---or \emph{facts}---using pattern matching.
Although such basic facts allow us to perform simple computational tasks,
such as keyword-based image classification and search, they are not enough
to understand the image.

To truly \emph{understand} what is being depicted in Figure~\ref{fig:marat}
we need more than basic facts.  We need (1)~general knowledge about the
world, (2)~specific knowledge about the persons named, and (3)~the capacity
to combine general and specific knowledge with the facts extracted from the
image in order to infer new facts.

Now suppose we are given~(1), (2), and (3).  From our general knowledge of
the world, and possibly by further analyzing the image, we can assert with
high confidence that Marat is holding the letter and that he has a stab
wound on the chest.  From this and from the blood-tainted knife depicted
below him, we might infer that the depicted knife is the object that caused
the wound.  Since knifes are not autonomous beings, we might also conclude
that someone (possibly himself) stabbed Marat in the chest.  But who and
why?

To answer these questions we will need more information.  Suppose we are
told that Marat was a journalist and political agitator, and one of the
leaders of a radical political faction in the Reign of Terror period of the
French Revolution (c.~1793).  Suppose we are also told that Charlotte
Corday, who signed the letter, was a declared political enemy of Marat---she
blamed him for a number of killings in Paris and other cities and believed
that he was a grave threat to the French Republic.  Under the light of these
new facts, we can conclude that Figure~\ref{fig:marat} looks like the scene
of a political murder.

By combining this conclusion with the additional fact that Charlotte Corday
is known to have murdered Jean-Paul Marat with a knife while he was in his
bathtub, holding a letter from her, we can infer that Figure~\ref{fig:marat}
must be a graphical representation of this incident, that is, of the
politically motivated assassination of Jean-Paul Marat by Charlotte Corday.

The derivation of this last fact from the visual patterns of
Figure~\ref{fig:marat} has only been possible because we have had access not
only to basic facts extracted from the image, but also to facts about the
world (common sense knowledge) and about the depicted objects and persons
(domain knowledge), and because we could combine all of these facts and make
inferences.

One of the main goals of symbolic AI is to enable the representation and
manipulation of pieces of knowledge by computers in ways that resemble or
emulate the kind of manipulations performed by humans---manipulations
similar to the considerations that enabled us to determine the true meaning
of Figure~\ref{fig:marat}.  The combination of this capacity with multimedia
opens up many possibilities.  For instance, the Marat's murder example is an
application of automated image understanding.  Two related applications are
video understanding and audio understanding, which are often more complex as
they involve the extraction of temporal information.

Other applications of symbolic AI to multimedia include the semantic
retrieval, classification, recommendation, and inspection of multimedia
data---for example, to automatically identify suspicious activity in
surveillance videos, generate age ratings for music and movies, and identify
risk factors for diseases in medical images and videos.  Although we could
list many other interesting applications, and we will discuss some of these
as we go along, here our focus is not on the applications themselves.  Our
focus is on the principles and technologies that enabled their conception in
the first place.

In this chapter, we give an introduction to symbolic AI from the perspective
of multimedia.  We begin, in Section~\ref{sec:DL}, with description logic
(DL).  One reason for doing so is that DL allows us to introduce symbolic
representation in an abstract setting, without the complications and biases
of a particular technology.  Another reason is that DL is itself a practical
technology: it is the logical underpinning of the most expressive family of
ontology languages, the OWL family.

After introducing DL, in the next section (Section~\ref{sec:semantic-web})
we discuss the incarnation of symbolic AI on the Web, the so-called
\emph{Semantic Web}.  We lay out the Semantic Web vision, its main enabling
technologies (RDF, OWL, SPARQL, SWRL, etc.), and the relation of these
technologies with the notions discussed in Section~\ref{sec:DL}.  Throughout
both sections, whenever possible we motivate the discussion with examples
from the multimedia domain.  We also strive to present or mention
methodologies and extensions specifically designed to solve the kinds of
problems that arise in multimedia representation.

We conclude the chapter with the discussion of a hybrid model for multimedia
representation (Section~\ref{sec:HK}).  This model, called
\emph{Hyperknowledge} (HK), generalizes a classic hypermedia model with the
notions of concept nodes and semantic links.  By doing so, the
Hyperknowledge model allows for the integrated representation and processing
of multimedia content together with their semantic description.

Suggestions for further reading are listed at the end of the chapter
(Section~\ref{sec:further-reading}).



\section{Description Logic(s)}
\label{sec:DL}

The word logic has many meanings and no completely satisfying definition.
We can understand logic as a system for deducing new statements from
previously asserted statements.  There are many kinds of logics.  One of the
simplest is propositional logic, which deals with statements constructed by
means of the propositional connectives $¬$ (not), $∧$ (and), $∨$ (or), $→$
(if-then), and $↔$ (if and only if).  First-order logic (FOL), often used to
formalize mathematical statements, extends propositional logic with the
notions of variables, constants, functions, predicates, and quantifiers.
There are still further extensions, such as second order logic, and logics
which adopt slightly different approaches, such as modal and fuzzy logics.

The term ``description logic'' (DL) refers not to a single logic but to a
\emph{family} of logics.  That is why we sometimes write description
logic\emph{s} (plural).  The logics in this family come in many flavors, but
most of them share the same core modeling features---they speak of
individuals, concepts, and roles.  Here we will focus on the description
logic \SROIQ~\cite{Horrocks-I-2006}.  Most of today's mainstream DLs are
sublanguages of \SROIQ, which is also the basis of the ontology language
OWL~2 DL~\cite{W3C-OWL2Overview}.  We will follow the usual approach for
presenting a formal language.  We begin by giving the form (syntax) of
\SROIQ\ statements together with their intuitive meaning.  We will define
their exact meaning (semantics) later.

\subsection{Syntax}
\label{sub:DL:syntax}

We assume we are given three sets of names, $𝖭_I$, $𝖭_C$, and $𝖭_R$,
containing names for individuals, concepts, and roles, respectively.  The
idea is that individuals will represent real-world entities, concepts will
represent characteristics of these entities, and roles will represent
relations between (two) entities.\footnotemark\ We assume further that the
sets $𝖭_I$, $𝖭_C$, and $𝖭_R$ are pairwise disjoint, that is, that no name
occurs at the same time in more than one of the sets.  The three sets of
names constitute the \emph{vocabulary} of \SROIQ---the basic building blocks
which we will use to construct statements of the language.

\footnotetext{For those versed in first-order logic, names of individuals
  are constants, names of concepts are unary predicates, and names of roles
  are binary predicates.}

We distinguish three kinds of statements, which in DL are also called
\emph{axioms}.  These are grouped into \emph{boxes}: ABox, TBox, and RBox.
The latter, RBox, is available only in very expressive description logics,
such as \SROIQ, but all DL-based ontologies have a TBox and most of them an
ABox.  Together the three boxes comprise a \emph{knowledge base}.

The ABox contains the \emph{assertional knowledge}: statements that assert
facts about specific individuals.  For example, the concept assertion
\[
  \TT{Director}(\T{kubrick})
\]
states that the individual named $\T{kubrick}$ is a director.  Similarly,
the role assertion
\[
  \TT{directs}(\T{kubrick},\TT{space-odyssey})
\]
states that the individual named $\T{kubrick}$ directs the individual (that
is, the movie) named $\TT{space-odyssey}$.

The TBox and RBox contain the \emph{terminological knowledge}: statements
that assert facts concerning all individuals.  The TBox contains statements
about concepts and the RBox contains statements about roles.
For example, the TBox axiom
\[
  \TT{Director}⊑\TT{Person}
\]
states that the concept named $\TT{Director}$ is subsumed by the concept
named $\TT{Person}$, that is, that every individual who is a director is
also a person.
And the RBox axiom
\[
  \TT{directs}⊑\TT{likes}
\]
states that the role named $\TT{directs}$ is subsumed by the role named
$\TT{likes}$, that is, that directing a movie implies liking
it.\footnotemark

\footnotetext{We can rewrite the last two statements in first-order logic as
  \[
    ∀x(\TT{Director}(x)→\TT{Person}(x))
    \quad\text{and}\quad
    ∀x∀y(\TT{directs}(x,y)→\TT{likes}(x,y)).
  \]
  The notation of DL is inspired by set theory and does not use variables.}

This last statement is, of course, questionable.  Maybe some director out
there in the world does not like some movie which he or she directed.  But
the point here is that the statements in a knowledge base, especially those
in the TBox and RBox, formalize a particular \emph{conceptualization} of a
domain.  The conceptualization reflects the world-view of its designer and
may contain simplifications which make sense in its context of use.

Another property of a conceptualization is that it is
language-independent---it is just a conceptual model which can be created in
a modeling language such as UML.  The term \emph{ontology} is sometimes used
to mean the instantiation of a conceptualization in a specific knowledge
representation language.  For our purposes, the term ontology will mean the
same as knowledge base, that is, a set of axioms partitioned into ABox,
TBox, and RBox.

Back to \SROIQ, we will now define the precise syntax of the statements in
each of the boxes.

\paragraph*{RBox}

The \SROIQ\ RBox contains statements that describe characteristics of roles
and interdependencies between roles.  A \emph{role} is one of the following
expressions:
\begin{enumerate}
\item A role name~$\T{r}$, for some $\T{r}$ in $𝖭_R$.
\item An inverted role name $\T{r}^-$, for some $\T{r}$ in $𝖭_R$.
\item The universal role~$u$.
\end{enumerate}

A \emph{role inclusion axiom} (RIA) is a statement of the form
\[
  r_1∘⋯∘r_n⊑r,
\]
where $r_1$, $…$, $r_n$, and $r$ are roles.

Here is an example RIA:
\[
  \TT{fatherOf}⊑\TT{childOf}^-.
\]
This RIA states that the role named $\TT{fatherOf}$ is subsumed by the
inverse of the role named $\TT{childOf}$, that is, that if $a$ is a father
of $b$ then $b$ is a child of $a$, for any individuals $a$ and $b$.
Similarly, the RIAs
\[
  \TT{owns}⊑\TT{caresFor}
  \quad\text{and}\quad
  \TT{owns}\mathbin{∘}\TT{partOf}⊑\TT{owns}
\]
state, respectively, that if $a$ owns $b$ then $a$ cares for $b$ (ownership
implies care) and that if $a$ owns $b$ and $b$ is part of $c$ then $a$ owns
$c$ (ownership of a part implies ownership of the whole).\footnotemark\ A
finite set of RIAs is called a \emph{role hierarchy}.

\footnotetext{These three RIAs can be formalized in first-order logic as
  \begin{gather*}
    ∀x∀y(\TT{fatherOf}(x,y)→\TT{childOf}(y,x))\\
    ∀x∀y(\TT{owns}(x,y)→\TT{caresFor}(x,y))\\
    ∀x∀y∀z(\TT{owns}(x,y)∧\TT{partOf}(y,z)→\TT{owns}(x,z)).
  \end{gather*}
}

A \emph{role characteristic} is a statement of one of the forms:
\[
  \SS{Sym}(r),\quad
  \SS{Asy}(r),\quad
  \SS{Tra}(r),\quad
  \SS{Ref}(r),\quad
  \SS{Irr}(r),\quad
  \SS{Dis}(r,s),
\]
where $r$ and $s$ are any roles different from the universal role $u$.  As
we will see in Section~\ref{sub:DL:semantics}, we interpret roles as binary
relations and we use role characteristic statements to assert properties of
these relations---namely, \emph{sym}metry, \emph{asy}mmetry,
\emph{tra}nsitivity, \emph{ref}lexivity, \emph{irr}eflexivity, and
\emph{dis}jointness.  We will give the precise meaning of these terms later.

A \SROIQ\ \emph{RBox} consists of a role hierarchy together with a finite
set of role characteristics.  The RBox is said to be \emph{regular} if its
role hierarchy is regular.

Regularity is desirable property because it guarantees the decidability of
the resulting logic.  In practice, this means that any reasoning task will
terminate in a finite time (not loop indefinitely).  We will not define what
a regular RBox is, as that would take us a little too far afield, but we
remark that regularity is a purely syntactical property: it can be obtained
by restricting the form of the RIAs that occur in a given role hierarchy.
See~\cite{Horrocks-I-2006} for details.

\paragraph*{TBox}

The \SROIQ\ TBox contains statements relating concepts, or more precisely,
concept expressions.  A \emph{concept (expression)} is one of the following:
\begin{enumerate}
\item\label{Cexp:1} A concept name $\T{C}$, for some $\T{C}$ in $𝖭_C$.
\item\label{Cexp:2} The \emph{top} concept $⊤$.
\item\label{Cexp:3} The \emph{bottom} concept $⊥$.
\item\label{Cexp:4} The \emph{nominal} concept $\{\T{a}_1,…,\T{a}_n\}$ where
  $\T{a}_1$, …, $\T{a}_n$ are names in $𝖭_I$.
\item\label{Cexp:5} The \emph{negation} $¬C$ where $C$ is a concept expression.
\item\label{Cexp:6} The \emph{intersection} $C⊓D$ where $C$ and $D$ are
  concept expressions.
\item\label{Cexp:7} The \emph{union} $C⊔D$ where $C$ and $D$ are concept
  expressions.
\item\label{Cexp:8} The \emph{existential} $∃{r}.C$ where $r$ is a role and
  $C$ is a concept expression.
\item\label{Cexp:9} The \emph{universal} $∀{r}.C$ where $r$ is a role and
  $C$ is a concept expression.
\item\label{Cexp:10} The \emph{self-restriction} $∃r.\S{Self}$ where $r$ is
  a (simple) role.
\item\label{Cexp:11} The \emph{at-least restriction} ${≥}nr.C$ where $n$ is
  a nonnegative integer, $r$ is a (simple) role, and $C$ is a concept
  expression.
\item\label{Cexp:12} The \emph{at-most restriction} ${≤}nr.C$ where $n$ is a
  nonnegative integer, $r$ is a (simple) role, and $C$ is a concept
  expression.
\end{enumerate}

Note that a concept expression, as defined above, is \emph{not} a
statement---it does not assert something which is either true or false.
Instead, a concept expression constructs a new concept from the given
expressions, which may be other concept expressions, individual names, or
roles.  This construction starts with the atomic concepts, that is, the
concept names and the top $⊤$ and bottom $⊥$ concepts (rules~\ref{Cexp:1},
\ref{Cexp:2}, and~\ref{Cexp:3} above).

Recall that a concept stands for a given characteristic of certain
individuals.  We can understand a concept $C$ as the collection (or
\emph{set}) of all individuals possessing that given characteristic.  So,
the concept name $\T{Actor}$ can be understood as the set of all individuals
who are actors.  Similarly, the top concept $⊤$ can be understood as the set
of all individuals whatsoever and the bottom concept $⊥$ as the empty
set---the set containing no individuals at all.

A nominal concept $\{\T{a}_1,…,\T{a}_n\}$ (rule~\ref{Cexp:4}) stands for the
set containing precisely the individuals named by $\T{a}_1$, $…$, $\T{a}_n$.
For instance, $\{\T{kubrick},\T{scorsese}\}$ stands for the set whose only
members are the individuals named $\T{kubrick}$ and $\T{scorsese}$.

The negation, intersection, and union operations (rules~\ref{Cexp:5},
\ref{Cexp:6}, and~\ref{Cexp:7}) allows us to construct new concepts from
other concepts.  For instance, the concept $¬\T{Actor}$ stands for the set
of all individuals who are not actors.  Similarly, the concepts
\[
  \T{Actor}⊓\T{Female}
  \quad\text{and}\quad
  \T{Actor}⊔\T{Director}
\]
stand, respectively, for the set of individuals who are female actors and
the set of individuals who are actors or directors (or both).  Note that
these operations apply to any concept expressions, not just names.  So, for
example, the complex concept
\[
  \T{Director}⊓¬(\T{Female}⊔\{\T{kubrick},\T{scorsese}\})
\]
stands for directors who are neither female nor the individuals named
$\T{kubrick}$ or $\T{scorsese}$.

The existential and universal quantifiers (rules~\ref{Cexp:8}
and~\ref{Cexp:9}) take a role and a concept and construct a new concept.
For instance, the concepts
\[
  ∃\T{knows}.\T{Actor}
  \quad\text{and}\quad
  ∀\T{directs}.\{\TT{a-bronx-tale,good-shepherd}\}
\]
stand, respectively, for the set individuals who know \emph{some} actor, and
for the set individuals who either haven't directed any film or directed
exactly the films named $\TT{a-bronx-tale}$ and $\TT{good-shepherd}$.  As we
will see in the next section, by the way the $∀$ operator is defined, if we
want to exclude the ``haven't directed any film part'' we must write
\[
  (∃\T{directs}.⊤)⊓(∀\T{directs}.\{\TT{a-bronx-tale,good-shepherd}\}),
\]
which incidentally specifies the set $\{\TT{de-niro}\}$.

The self-restriction operator (rule~\ref{Cexp:10}) takes a (simple) role $r$
and constructs the concept containing all individuals which are related to
themselves via $r$.  For example, the concept $∃\T{loves}.\S{Self}$ stands
for the set of all individuals who love themselves.  Clearly, this operator
only makes sense when applied to roles which connect the same kind of
entity, for example, persons to persons, movies to movies, etc.

The requirement of $r$ being \emph{simple} in the definition of the
self-restriction operator is related to the problem of regularity mentioned
earlier.  The roles that appear in an \SROIQ\ RBox are classified into
simple and non-simple depending on the form of the RIAs in which they occur.
Informally, the non-simple roles are those which are ``created'' by role
chains of length greater than one; the remaining roles are considered
simple.  We assume that the $r$ and $s$ occurring in role characteristics of
the form $\SS{Irr}(r)$, $\SS{Asy}(r)$, and $\SS{Dis}(r,s)$ are always simple
roles.  See~\cite{Horrocks-I-2006} for details.

The two remaining concept constructors of \SROIQ\ are the cardinality
restriction operators $≥$ and $≤$ (rules~\ref{Cexp:11} and \ref{Cexp:12}).
Both take a nonnegative integer $n$, a simple role $r$, and concept $C$, and
construct a new concept containing all individuals which are $r$-connected
to at-least or at-most $n$ individuals in $C$.  For example, the concepts
\[
  {≥}2\T{knows}.\T{Actor}
  \quad\text{and}\quad
  {≤}5\T{directs}.⊤
\]
stand, respectively, for the set of individuals who know two or more actors,
and the set of individuals who directed five or less movies.

We are finally in a position to define what a \SROIQ\ TBox is.  The
\emph{\SROIQ\ TBox} is a finite set of \emph{general concept inclusion
  axioms} (GCIs) where each GCI is a statement of the form
\[
  C⊑D,
\]
for arbitrary concepts $C$ and $D$.  As in RIAs, the symbol $⊑$ in GCIs
stands for subsumption.  So, for instance,
\[
  \T{Actor}⊑\T{Person}
  \quad\text{and}\quad
  ∃\T{directs}.⊤⊑∃\T{knows}.\T{Actor}
\]
state, respectively, that the every individual who is an actor is also a
person, and that every individual who directs some film knows some
actor.\footnotemark

\footnotetext{To translate a GCI to first-order logic we first translate the
  concepts on each side of the $⊑$ as formulas with a single free variable,
  say $x$.  The $⊑$ becomes an implication and the whole formula is then
  enclosed in a $∀x$.  For example, the last two GCIs are translated as:
  \[
    ∀x(\T{Actor}(x)→\T{Person}(x))
    \quad\text{and}\quad
    ∀x(∃y(\T{directs}(x,y)∧⊤(y))→∃z(\T{knows}(x,z)∧\T{Actor}(z))).
  \]
}

\paragraph*{ABox}

The \SROIQ\ ABox contains the \emph{assertional statements}: statements that
assert facts about specific individuals.  These statements have one of the
forms:
\[
  C(\T{a}),\quad
  r(\T{a},\T{b}),\quad
  ¬r(\T{a},\T{b}),\quad
  \T{a}≈\T{b},\quad
  \T{a}≉\T{b},
\]
where $C$ is a concept, $r$ is a role, and $\T{a}$ and $\T{b}$ are names of
individuals.

The \emph{concept assertion} $\TT{Actor}⊓\TT{Director}(\TT{de-niro})$ states
that the individual named $\TT{de-niro}$ is an \emph{instance} of the
concept named $\TT{Actor}⊓\TT{Director}$, that is, that this individual is
both an actor and a director.

The \emph{role assertion} $\TT{directs}(\TT{kubrick},\TT{dr-strangelove})$
states that the individual named $\TT{kubrick}$ is $\TT{directs}$-connected
to the individual $\TT{dr-strangelove}$.  Similarly, the \emph{negated role
  assertion} $¬\TT{directs}(\TT{kubrick},\TT{taxi-driver})$ states that
$\TT{kubrick}$ is not $\TT{directs}$-connected to $\TT{taxi-driver}$.

The \emph{equality assertion} $\TT{kubrick}≈\TT{stanley}$ and the
\emph{inequality assertion} $\TT{kubrick}≉\TT{taxi-driver}$ state,
respectively, that $\TT{kubrick}$ and $\TT{stanley}$ name the same
individual, and that $\TT{kubrick}$ and $\TT{taxi-driver}$ do not name the
same individual.

\paragraph*{Knowledge base (KB)}

A \emph{\SROIQ\ knowledge base} is the union of the three boxes: RBox, TBox,
and ABox.  A knowledge base is said to be regular if its RBox is regular.

\paragraph*{The movie facts KB}

We conclude this section on the syntax of \SROIQ\ with an example.  The
knowledge base below collects some of the \emph{movie facts} we have been
discussing.  Note that the usual interpretation of these statements---the
interpretation in which $\T{kubrick}$ names the famous director---is just
one of many possible interpretations.  We will have more to say about
interpretations next.

\begingroup
  \small
  \newcommand*\X{\rule{0pt}{14pt}}
  \newcommand*\A[1]{{\footnotesize``#1''}}
  \setlength\LTleft{0pt}
  \setlength\LTright{0pt}
  \begin{longtable}{|l@{\extracolsep{\fill}}r|}
    \hline\multicolumn{2}{|l|}{\X\textsl{ABox}}\\[1\jot]
    $\T{Director}(\T{kubrick})$
    &\A{$\T{kubrick}$ is a director}\\
    $\T{Actor}⊓\T{Director}(\TT{de-niro})$
    &\A{$\TT{de-niro}$ is an actor and a director}\\
    $\T{{≥}2knows.Actor}(\TT{stanley})$
    &\A{$\TT{stanley}$ knows at least two actors}\\
    $\TT{directs}(\T{kubrick},\TT{space-odyssey})$
    &\A{$\T{kubrick}$ directed $\TT{space-odyssey}$}\\
    $¬\TT{directs}(\T{kubrick},\TT{taxi-driver})$
    &\A{$\T{kubrick}$ did not direct $\TT{taxi-driver}$}\\
    $\T{kubrick}≈\T{stanley}$
    &\A{$\T{kubrick}$ and $\TT{stanley}$ are the same individual}\\
    $\T{kubrick}≉\TT{de-niro}$
    &\A{$\T{kubrick}$ and $\TT{de-niro}$ are not the same individual}\\
    \hline\multicolumn{2}{|l|}{\X\textsl{TBox}}\\[1\jot]
    $\TT{Director}⊑∃\T{directs}.⊤$
    &\A{a director directs something}\\
    $∃\T{directs}.⊤⊑\T{Director}$
    &\A{anything which directs is a director}\\
    $\T{Director}⊑\T{Person}$
    &\A{a director is a person}\\
    $\T{Director}⊑{≥}1\T{knows}.\T{Actor}$
    &\A{a director knows at least one actor}\\
    \multicolumn{2}{|l|}{
      $(∃\T{directs}.⊤)⊓(∀\T{directs}.\{\TT{a-bronx-tale,good-shepherd}\})
      ⊑\{\TT{de-niro}\}$
    }\\
    \multicolumn{2}{|r|}{
      \A{anything which directs exactly $\{\TT{a-bronx-tale,good-shepherd}\}$
        must be a $\{\TT{de-niro}\}$}
    }\\
    \hline\multicolumn{2}{|l|}{\X\textsl{RBox}}\\[1\jot]
    $\T{directs}⊑\T{likes}$
    &\A{directing implies liking}\\
    $\T{directs}∘\T{actsIn}^-⊑\T{knows}$
    &\A{if $x$ directs $y$ in which $z$ acts, then $x$ knows $z$}\\
    \hline
  \end{longtable}
\endgroup

\subsection{Semantics}
\label{sub:DL:semantics}

A knowledge base, as defined previously, is basically a set well-formed
strings, called statements or axioms.  The meaning of each of these strings
depends ultimately on the meaning of the names which occur in them.  To see
this, consider the axiom $\T{Director}(\T{kubrick})$.  This axiom will be
true if we interpret the names $\T{Director}$ and $\T{kubrick}$ as ``the
property of being a movie director'' and ``Stanley Kubrick'', respectively.
If however we interpret $\T{Director}$ as ``the property of being an even
number'' and $\T{kubrick}$ as ``the number 37'', then the assertion
$\T{Director}(\T{kubrick})$ ceases to be true.  This example is artificial
but serves to illustrate our point that the meaning is derived from the
interpretation of the names.  We will now define this notion of
interpretation precisely.

\paragraph*{Interpretation}

An \emph{interpretation} $\I$ consists of two things:
\begin{enumerate}
\item A nonempty set $Δ^\I$ called the \emph{domain} or \emph{universe} of
  discourse.
\item A function $⬚^\I$ from the vocabulary sets $𝖭_I$, $𝖭_C$, and $𝖭_R$
  such that:\footnotemark
  \begin{enumerate}
  \item If $\T{a}∈𝖭_I$ then $\T{a}^\I∈Δ^\I$.
  \item If $\T{C}∈𝖭_C$ then $\T{C}^\I⊆Δ^\I$.
  \item If $\T{r}∈N_R$ then $\T{r}^\I⊆Δ^\I×Δ^\I$.
  \end{enumerate}
\end{enumerate}

\footnotetext{Note that $x^\I$ is just a more compact way to write $\I(x)$
  which is the usual syntax for function application.}

In other words, an interpretation $\I$ fixes a domain of discourse $Δ^\I$,
which can be any nonempty set, and maps the names of individuals into
elements of the domain, the names of concepts into subsets of the domain,
and the names of roles into binary relations on the domain.

We reserve the term \emph{individual} to refer to the semantic counterpart
of an individual name, that is, to an element of the domain.  And we reserve
the terms \emph{concept extension} and \emph{role extension} to refer,
respectively, to the semantic counterparts of concepts and roles, that is,
to unary and binary relations on the domain.

We now lift the interpretation function $⬚^\I$ so that it can be applied to
arbitrary roles and concepts:
\begingroup
\interdisplaylinepenalty=100
\begin{align}
  u^\I
  &=Δ^\I×Δ^\I\label{Ilifted:1}\\
  (r^-)^\I
  &=\{⟨δ',δ⟩∣⟨δ,δ'⟩∈r^\I\}\label{Ilifted:2}\\
  ⊤^\I
  &=Δ^\I\label{Ilifted:3}\\
  ⊥^\I
  &=∅\label{Ilifted:4}\\
  \{\T{a}_1,…,\T{a}_n\}^\I
  &=\{\T{a}_1^\I,…,\T{a}_n^\I\}\label{Ilifted:5}\\
  (¬C)^\I
  &=Δ^\I⧵C^\I\label{Ilifted:6}\\
  (C⊓D)^\I
  &=C^\I∩D^\I\label{Ilifted:7}\\
  (C⊔D)^\I
  &=C^\I∪D^\I\label{Ilifted:8}\\
  (∃r.C)^\I
  &=\{δ∈Δ^\I∣\text{$⟨δ,δ'⟩∈r^\I$ and $δ'∈C^\I$, for some $δ'∈Δ^\I$}\}
    \label{Ilifted:9}\\
  (∀r.C)^\I
  &=\{δ∈Δ^\I∣\text{if $⟨δ,δ'⟩∈r^\I$ then $δ'∈C^\I$, for all $δ'∈Δ^\I$}\}
    \label{Ilifted:10}\\
  (∃r.\S{Self})^\I
  &=\{δ∈Δ^\I∣⟨δ,δ⟩∈r^\I\}\label{Ilifted:11}\\
  ({≥}nr.C)^\I
  &=\{δ∈Δ^\I∣\#\{δ'∈Δ^\I∣⟨δ,δ'⟩∈r^\I\text{\ and\ }δ'∈C\}≥n\}\label{Ilifted:12}\\
  ({≤}nr.C)^\I
  &=\{δ∈Δ^\I∣\#\{δ'∈Δ^\I∣⟨δ,δ'⟩∈r^\I\text{\ and\ }δ'∈C\}≤n\}\label{Ilifted:13}
\end{align}
\endgroup

These equations can be read as follows:
\begin{enumerate}
\item The \emph{universal role} $u$ stands for (denotes) the relation which
  connects every two individuals in the domain.
\item The \emph{inverted role} $r^-$ denotes the same relation as $r$ but
  with the first and second coordinates swapped.
\item The \emph{top concept} $⊤$ denotes the whole domain.
\item The \emph{bottom concept} $⊥$ denotes the empty set.
\item The \emph{nominal} $\{\T{a}_1,…,\T{a}_n\}$ denotes the set containing
  precisely the individuals denoted by the names $\T{a}_1$, $…$, $\T{a}_n$.
\item The \emph{negation} $¬C$ denotes the difference between the domain and
  the set denoted by $C$, that is, the set of all individuals in $Δ^\I$
  which are not in $C^\I$.
\item The \emph{intersection} $C⊓D$ denotes the intersection of the sets
  denoted by $C$ and $D$, that is, the set of all individuals which belong
  to both $C^\I$ and $D^\I$.
\item The \emph{union} $C⊔D$ denotes the union of the sets denoted by $C$
  and $D$, that is, the set of all individuals which belong to at least one
  of $C^\I$ or $D^\I$.
\item The \emph{existential} $∃r.C$ denotes the set of all individuals which
  are connected via relation $r^\I$ to some individual in $C^\I$.
\item The \emph{universal} $∀r.C$ denotes the set of all individuals $a$
  such that if $a$ is connected to some individual $b$ via $r^\I$ then $b$
  is in $C^\I$.  A consequence of this conditional form is that any
  individual which is not connected to another via $r^\I$ will be in the
  resulting set (as the implication is vacuously true).\footnotemark
\item The \emph{self restriction} $∃r.\S{Self}$ denotes the set of all
  individuals which are connected to themselves via relation $r^\I$.
\item The \emph{at-least restriction} ${≥}n.rC$ denotes the set of all
  individuals which are connected via relation $r^\I$ to at least $n$
  individuals in $C^\I$.
\item The \emph{at-most restriction} ${≤}n.rC$ denotes the set of all
  individuals which are connected via relation $r^\I$ to at most $n$
  individuals in $C^\I$.
\end{enumerate}

\footnotetext{%
  Recall that in classical logic $A→B$ is equivalent to $(¬A)∨B$.  So, we
  can restate equation~\eqref{Ilifted:10} as follows:
  \[
    (∀r.C)^\I
    =\{δ∈Δ^\I∣\text{$⟨δ,δ'⟩∉r^\I$ or $δ'∈C^\I$, for all $δ'∈Δ^\I$}\}.
  \]
  Clearly, any $δ$ which is not $r$-connected to some $δ'$ satisfies the
  above condition.}

We now turn to the problem of defining the truth-value (truth or falsity) of
an axiom under a given interpretation.

\paragraph*{Satisfiability}

An interpretation $\I$ \emph{satisfies} (or \emph{is a model of}) an axiom
$α$, in symbols $\I⊨α$, under the following conditions:
\begin{enumerate}
\item (RIA) $\I⊨r_1∘⋯∘r_n⊑r$ iff\footnote{We will write ``iff'' as an
    abbreviation for ``if and only if''.} for every sequence $δ_0$, $δ_1$,
  …, $δ_n$ in $Δ^\I$:
  \[
    ⟨δ_0,δ_1⟩∈r_1^\I,\
    ⟨δ_1,δ_2⟩∈r_2^\I,\
    …,
    \ \text{and}\
    ⟨δ_{n-1},δ_n⟩∈r_n^\I
    \quad\text{implies}\quad
    ⟨δ_{0},δ_n⟩∈r^\I,
  \]
  that is, iff every path in $Δ^\I$ that traverses $r_1^\I$, $…$, $r_n^\I$
  (in this order) has a direct $r^\I$-shortcut.
\item (Role symmetry) $\I⊨\SS{Sym}(r)$ iff $r^\I$ is a symmetric relation,
  that is, iff $⟨δ_1,δ_2⟩∈r^\I$ implies $⟨δ_2,δ_1⟩∈r^\I$ and vice versa, for
  any $δ_1,δ_2$ in $Δ^\I$.
\item (Role asymmetry) $\I⊨\SS{Asy}(r)$ iff $r^\I$ is an asymmetric
  relation, that is, iff $⟨δ_1,δ_2⟩∈r^\I$ implies $⟨δ_2,δ_1⟩∉r^\I$, for any
  $δ_1,δ_2$ in $Δ^\I$.
\item (Role transitivity) $\I⊨\SS{Tra}(r)$ iff $r^\I$ is a transitive
  relation, that is, iff $⟨δ_1,δ_2⟩∈r^\I$ and $⟨δ_2,δ_3⟩∈r^\I$ implies
  $⟨δ_1,δ_3⟩∈r^\I$, for any $δ_1,δ_2,δ_3$ in $Δ^\I$.
\item (Role reflexivity) $\I⊨\SS{Ref}(r)$ iff $r^\I$ is a reflexive
  relation, that is, iff $⟨δ,δ⟩∈r^\I$, for any $δ$ in $Δ^\I$.
\item (Role irreflexivity) $\I⊨\SS{Irr}(r)$ iff $r^\I$ is an irreflexive
  relation, that is, iff $⟨δ,δ⟩∉r^\I$, for any $δ$ in $Δ^\I$.
\item (Role disjointness) $\I⊨\SS{Dis}(r,s)$ iff $r^\I∩s^\I=∅$, that is, iff
  no two individuals are simultaneously $r^\I$-connected and
  $s^\I$-connected.
\item (GCI) $\I⊨C⊑D$ iff $C^\I⊆D^\I$, that is, iff every individual of
  $C^\I$ is also an individual of $D^\I$.
\item (Concept assertion) $\I⊨C(\T{a})$ iff $\T{a}^\I∈C^\I$, that is, if the
  individual denoted by $\T{a}$ belongs to the extension of concept $C$.
\item (Role assertion) $\I⊨r(\T{a},\T{b})$ iff $⟨\T{a}^\I,\T{b}^\I⟩∈r^\I$,
  that is, iff $\T{a}^\I$ is $r^\I$-connected to $\T{b}^\I$.
\item (Negated role assertion) $\I⊨¬r(\T{a},\T{b})$ iff $⟨\T{a}^\I,\T{b}^\I⟩∉r^\I$.
\item (Equality assertion) $\I⊨\T{a}≈\T{b}$ iff $\T{a}^\I=\T{b}^\I$, that
  is, iff $\T{a}^\I$ and $\T{b}^\I$ are the same element of $Δ^\I$.
\item (Inequality assertion) $\I⊨\T{a}≉\T{b}$ iff $\T{a}^\I≠\T{b}^\I$.
\end{enumerate}

Now that we have defined when an interpretation satisfies an axiom, we can
extend this notion to apply to a knowledge base (set of axioms).  An
interpretation $\I$ \emph{satisfies} (or \emph{is a model of}) of a
knowledge base $\KB$, in symbols $\I⊨\KB$, iff $\I⊨α$, for every axiom $α$
in $\KB$.  A knowledge base is said to be \emph{satisfiable} or
\emph{consistent} if it has a model, that is, if there is some
interpretation $\I$ which satisfies it; otherwise, the knowledge base is
said to be \emph{insatisfiable} or \emph{inconsistent}.

To make the discussion of interpretations more concrete, consider the
following interpretation $\I$ for the movie facts KB presented at the end of
the last section:
\begingroup
\vskip.5\abovedisplayskip
\noindent
\begin{tikzpicture}[remember picture]
  \node[text width=.65\textwidth](A){%
    \abovedisplayskip=0pt
    \belowdisplayskip=0pt
    \begin{align*}
      Δ^\I&=\{i_1,i_2,i_3,i_4,i_5,i_6,i_7\}\\
      \T{kubrick}^\I&=i_1\\
      \TT{de-niro}^\I&=i_2\\
      \TT{stanley}^\I&=i_1\\
      \TT{space-odyssey}^\I&=i_3\\
      \TT{taxi-driver}^\I&=i_4\\
      \TT{a-bronx-tale}^\I&=i_5\\
      \TT{good-shepherd}^\I&=i_6\\
      \TT{Director}^\I&=\{i_1,i_2\}\\
      \TT{Actor}^\I&=\{i_2,i_7\}\\
      \TT{Person}^\I&=\{i_1,i_2\}\\
      \TT{knows}^\I&=\{⟨i_1,i_2⟩,⟨i_1,i_7⟩,⟨i_2,i_7⟩\}\\
      \TT{directs}^\I&=\{⟨i_1,i_3⟩,⟨i_2,i_5⟩,⟨i_2,i_6⟩\}\\
      \TT{likes}^\I&=\{⟨i_1,i_3⟩,⟨i_1,i_4⟩,⟨i_2,i_5⟩,⟨i_2,i_6⟩\}\\
      \TT{actsIn}^\I&=\{⟨i_7,i_5⟩\}
    \end{align*}};
  \node[anchor=north west](knows)
  at (A.north east){\strut$\TT{knows}^\I$};
  \node[anchor=north west](directs)
  at ($(A.north east)!.25!(A.south east)$){\strut$\TT{directs}^\I$};
  \node[anchor=north west](likes)
  at ($(A.north east)!.5!(A.south east)$){\strut$\TT{likes}^\I$};
  \node[anchor=north west](actsIn)
  at ($(A.north east)!.75!(A.south east)$){\strut$\TT{actsIn}^\I$};
  \begin{scope}[local bounding box=knowsbb]
    \draw (knows.north west)
    rectangle ($({directs.north east-|current page text area.east})+(0,.5)$);
  \end{scope}
  \begin{scope}[local bounding box=directsbb]
    \draw (directs.north west)
    rectangle ($({likes.north east-|current page text area.east})+(0,.5)$);
  \end{scope}
  \begin{scope}[local bounding box=likesbb]
    \draw (likes.north west)
    rectangle ($({actsIn.north east-|current page text area.east})+(0,.5)$);
  \end{scope}
  \begin{scope}[local bounding box=actsInbb]
    \draw (actsIn.north west)
    rectangle ($({A.south east-|current page text area.east})+(0,.5)$);
  \end{scope}
  \begin{scope}[node distance=1.2em,inner sep=2pt]
    \begin{scope}
      \node(k2) at ($(knowsbb)-(0,.25)$){$i_2$};
      \node[left=of k2](k1){$i_1$};
      \node[right=of k2](k7){$i_7$};
      \draw[->](k1) to (k2);
      \draw[->,bend left=35](k1) to (k7);
      \draw[->](k2) to (k7);
    \end{scope}
    \begin{scope}
      \node(d2) at ($(directsbb)+(0,.5)$){$i_2$};
      \node[left=of d2](d1){$i_1$};
      \node[below=of d2,yshift=.5em](d3){$i_3$};
      \node[right=of d2](d5){$i_5$};
      \node[below=of d5,yshift=.5em](d6){$i_6$};
      \draw[->](d1) to (d3);
      \draw[->](d2) to (d5);
      \draw[->](d2) to (d6);
    \end{scope}
    \begin{scope}
      \node(l2) at ($(likesbb)+(0,.5)$){$i_2$};
      \node[left=of l2](l1){$i_1$};
      \node[below=of l1,yshift=.5em](l4){$i_4$};
      \node[below=of l2,yshift=.5em](l3){$i_3$};
      \node[right=of l2](l5){$i_5$};
      \node[below=of l5,yshift=.5em](l6){$i_6$};
      \draw[->](l1) to (l3);
      \draw[->](l1) to (l4);
      \draw[->](l2) to (l5);
      \draw[->](l2) to (l6);
    \end{scope}
    \begin{scope}
      \node(a7) at ($(actsInbb)-(1,.25)$){$i_7$};
      \node[right=of a7](a5){$i_5$};
      \draw[->](a7) to (a5);
    \end{scope}
  \end{scope}
\end{tikzpicture}
\vskip.5\belowdisplayskip
\endgroup
\noindent The domain $Δ^\I$ in this case consists of seven distinct
individuals, $i_1$, $…$, $i_7$, and the names that occur in the movie facts
KB are mapped as shown above.  The following remarks are in order:
\begin{itemize}
\item Some individuals in the domain $Δ^\I$ might not have corresponding
  names: $i_7$ is not named by any name in the movie facts KB.
\item Two distinct names might denote the same thing: $\T{kubrick}$ and
  $\T{stanley}$ denote the individual $i_1$.
\item Concept names are mapped into subsets of the domain $Δ^\I$:
  $\T{Director}$, $\T{Actor}$, and $\T{Person}$ denote sets of individuals.
\item Role names are mapped into binary relations on the domain $Δ^\I$:
  $\T{knows}$, $\T{directs}$, $\T{likes}$, and $\T{actsIn}$ denote sets
  of ordered pairs of individuals.  The individuals $i_2$ and $i_5$ stand in
  relation $\T{directs}^\I$ to each other (in this other) because the pair
  $⟨i_2,i_5⟩$ is in $\T{directs}^\I$.  Binary relations can be conveniently
  depicted as directed graphs.  The graphs corresponding to the relations
  $\T{knows}^\I$, $\T{directs}^\I$, $\T{likes}^\I$, and $\T{actsIn}^\I$
  are shown in the boxes above.
\end{itemize}

Let $\KB$ stands for the movie facts KB and let $\I$ stands for the previous
interpretation.  The natural question to ask is whether $\I⊨\KB$, that is,
whether the previous interpretation satisfies the movie facts KB.  This will
be the case only if $\I⊨α$ for every axiom $α$ in the movie facts KB.  Let
us check that.

We begin by checking the movie facts ABox.  Clearly,
\[
  \I⊨\TT{Director}(\T{kubrick})
\]
because $\T{kubrick}^\I∈\TT{Director}^\I$, that is, because
$i_1∈\{i_1,i_2\}$, as required by the definition of satisfiability of
concept assertions.  Similarly,
\[
  \I⊨¬\T{directs}(\T{kubrick},\TT{taxi-driver})
\]
because $⟨\T{kubrick}^\I,\TT{taxi-driver}^\I⟩∉\T{directs}^\I$.  The
remaining axioms in the ABox are also satisfied by $\I$, as the reader can
easily check.

We will now check the movie facts TBox.  We have that
\[
  \I⊨\T{Director}⊑\T{Person}
\]
since every individual in the set $\T{Director}^\I$ is also in the set
$\T{Person}^\I$.  It is also the case that
\[
  \I⊨(∃\T{directs}.⊤)⊓(∀\T{directs}.\{\TT{a-bronx-tale,good-shepherd}\})
  ⊑\{\TT{de-niro}\}
\]
because:
\begin{enumerate}
\item $\TT{de-niro}^\I$ is connected via relation $\TT{directs}^\I$ to
some member of $⊤^\I$ (that is, the whole domain).
\item Everything to which $\TT{de-niro}^\I$ is $\TT{directs}^\I$-connected
  is an element of the set $\{\TT{a-bronx-tale}^\I,\TT{good-shepherd}^\I\}$.
\item The previous assertions (1) and (2) hold for no other individual
  besides $\TT{de-niro}^\I$.
\end{enumerate}
The rest of the axioms in the movie facts TBox are also satisfied by $\I$.
(Again, we leave these checks to the reader.)

The last thing to check are the axioms in the movie facts RBox.  We have
that
\[
  \I⊨\T{directs}⊑\T{likes}
\]
holds because every pair in relation $\T{directs}^\I$ is also in relation
$\T{likes}^\I$.  One way to determine whether or not
\[
  \I⊨\T{directs}∘\T{actsIn}^-⊑\T{knows}
\]
holds is to search for three (not necessarily distinct) individuals $x$,
$y$, $z$ such that $⟨x,y⟩∈\T{directs}^\I$ and $⟨z,y⟩∈\T{actsIn}^\I$ but
$⟨x,z⟩∉\T{knows}^\I$.  If we can find such three individuals then the
assertion is false (as we just found a counterexample).  Otherwise, if we
cannot find such three individuals, the assertion is true.  In the case of
the interpretation $\I$, there is only one possible choice of individuals
that satisfies the first two requirements, namely, $x=i_2$, $y=i_5$, and
$z=i_7$.  Under this choice the third requirement is also satisfied since
$⟨i_2,i_7⟩∈\T{knows}^\I$.  Hence the assertion is true, that is, the axiom
$\T{directs}∘\T{actsIn}^-⊑\T{knows}$ is satisfied by the interpretation
$\I$.

Since $\I$ satisfies all axioms in each of the boxes of the movie facts KB,
we conclude that answer to our original question (namely, whether or not
$\I⊨\KB$) is positive.  That is, it is indeed the case that the
interpretation $\I$ satisfies the movie facts KB, and consequently, the
movie facts KB is consistent (satisfiable).

Two related questions concerning the movie facts KB are the following: Can
we find an interpretation $\J$ that does not satisfies the movie facts KB?
And how can we modify the movie facts KB so that it becomes inconsistent
(insatisfiable)?

A simple answer to the first question is to take an interpretation $\J$
which is identical to the previous interpretation $\I$ but in which we map
$\T{kubrick}^\J$ and $\T{stanley}^\J$ to distinct individuals in the domain
$Δ^J$.  Clearly, under such interpretation the ABox axiom
$\T{kubrick}≈\T{stanley}$ does not hold, and so $\J⊭\KB$.

Concerning the second question, in order to make the movie facts KB
inconsistent we need to modify it in a way that prevents the construction of
any satisfying interpretation.  A trivial way to accomplish this is to add
the inequality assertion $\T{kubrick}≉\T{kubrick}$ to the KB.  This axiom
will be false under any interpretation $\I$ because whatever the choice of
denotation for $\T{kubrick}$ we will always have
$\T{kubrick}^\I=\T{kubrick}^\I$.  Of course, there are less trivial ways to
obtain an inconsistent KB.  (Can the reader find one involving a TBox
axiom?)

This concludes our discussion of satisfiability.  The whole point of a
formal semantics is to define a \emph{consequence relation} which allows us
to determine when a knowledge base entails a given axiom.  We will now
define this consequence relation in terms of the satisfiability relation.

\paragraph*{Logical consequence}

An axiom $α$ is a \emph{logical consequence} of (or \emph{is entailed by}) a
knowledge base $\KB$, in symbols $\KB⊨α$, iff $\I⊨\KB$ implies $\I⊨α$, for
every interpretation~$\I$.  That is, iff every interpretation which is a
model of the knowledge base $\KB$ is also a model of the axiom~$α$.\footnotemark

\footnotetext{Note that the symbol $⊨$ has different meanings depending on
  the kind of the object that occurs on its left-hand side.  When this
  object is interpretation, like in $\I⊨α$, the symbol $⊨$ stands for the
  satisfiability relation.  When this object is a set of axioms, like in
  $\KB⊨α$, the symbol $⊨$ denotes the logical consequence relation.}

Using this definition of logical consequence we can determine precisely when
an axiom $α$ follows from a given set of axioms.  As we discussed
previously, some KBs are inconsistent in the sense that there is no
interpretation satisfying all their axioms.  The problem with having an
inconsistent knowledge base is that, by definition, it entails any axiom.
That is, if $\KB$ is inconsistent then the assertion ``$\I⊨\KB$ implies
$\I⊨α$ for any $\I$'' is vacuously true because the antecedent of the
implication, namely, $\I⊨\KB$, is false.  This amounts to the saying that
states that anything follows from a contradiction. In practice,
inconsistency is an indicator of modeling errors.

An important task in DL theory is thus to determine whether or not a given
KB is consistent.  We will discuss this and other so-called \emph{reasoning
  tasks} in detail in the next section.  Before that, however, let us end
this section with an example of logical consequence.

Consider the following assertion about the movie facts $\KB$:
\[
  \KB⊨\TT{Actor}⊑\TT{Person}.
\]
This assertion states that the axiom $\TT{Actor}⊑\TT{Person}$ is logical
consequence of the set of axioms which comprise the $\KB$.  Is this
assertion true?  How can we prove it or disprove it?

Let us try to disprove it.  What we need to do in this case is find an
interpretation $\I$ which satisfies all axioms in $\KB$ but which fails to
satisfy $\TT{Actor}⊑\TT{Person}$.  For instance, take the previously
discussed interpretation $\I$ for $\KB$ whose domain consists of the
individuals $i_1$, $…$, $i_7$.  As we have checked a few paragraphs ago,
$\I⊨\KB$.  But it is not the case that every actor is a person under $\I$.
For instance, $i_7∈\TT{Actor}^\I$ but $i_7∉\TT{Person}^\I$.  Hence
$\KB⊭\TT{Actor}⊑\TT{Person}$, that is, the movie facts $\KB$ does not entail
the axiom $\TT{Actor}⊑\TT{Person}$.

It is usually harder to prove the opposite, that is, that a knowledge base
entails a given axiom $α$.  This amounts to showing that any attempt to find
an interpretation which satisfies the knowledge base but falsifies $α$ is
doomed to failure.  For instance, let us prove that the movie facts $\KB$
entails the axiom
\begin{equation*}
  ∃\T{likes}.⊤(\TT{de-niro}).
\end{equation*}
In other words, let us prove that that once we assume the axioms in the
movie facts $\KB$ are true we must conclude, necessarily, that
$\TT{de-niro}$ likes something.

Suppose there is an interpretation $\J$ which satisfies the movie facts
$\KB$ but which does not satisfy the above axiom.  From the assumption that
$\J⊨\KB$, we can infer that
\begin{align}
  \J&⊨\T{Actor}⊓\T{Director}(\TT{de-niro})\label{entails:1}\\
  \J&⊨\T{Director}⊑∃\T{directs}.⊤\label{entails:2}\\
  \J&⊨\T{directs}⊑\T{likes}\label{entails:3}
\end{align}
By~\eqref{entails:1}, we know that the individual denoted by $\TT{de-niro}$
must be in the set $\T{Director}^\J$.  And so, by~\eqref{entails:2}, we know
that this individual must be $\T{directs}^\J$-connected to someone, that is,
$⟨\TT{de-niro}^\J,a⟩∈\T{directs}^\J$, for some individual $a$.  Finally,
by~\eqref{entails:3}, it must be the case that
$⟨\TT{de-niro},a⟩∈\T{likes}^\J$, which contradicts our original assumption
that $\J⊭∃\T{likes}.⊤(\TT{de-niro})$.  Hence, we must conclude that no such
$\J$ can exist.  That is, any interpretation which satisfies the movie facts
$\KB$ will also satisfy $∃\T{likes}.⊤(\TT{de-niro})$, and so
$\KB⊨∃\T{likes}.⊤(\TT{de-niro})$.

\subsection{Reasoning tasks}
\label{sub:DL:reasoning}

One of the advantages of formalizing a body of knowledge in logic is the
possibility of treating it as a tangible object to which certain operations
can be applied.  In DL, the object is the KB (set of axioms) and the
operations are \emph{reasoning tasks} which attempt to extract new knowledge
from it.

An important characteristic of the reasoning tasks of DL, which make them at
same time powerful and complex, is that they follow the \emph{open-world
  assumption} (OWA).  Under this assumption, facts which cannot be deduced
from a KB are assumed to be \emph{unknown} but not necessarily false.  This
contrasts with the \emph{closed-world assumption} (CWA) often adopted in
database systems where facts that are not in the database are assumed to be
false.  That said, when needed the CWA can be emulated to some extent in DL
via nominals, and there specialized (autoepistemic, circumscriptive) DLs
with support for the CWA.

We will now describe briefly the principal kinds of reasoning tasks for DL.
For a more detailed description (with references)
see~\cite{Rudolph-S-2011,Baader-F-2007}.  A thorough discussion of the
computational complexities associated with these tasks for specifics DLs can
be found in~\cite{Complexity}, and a list of software implementing some of
the tasks can be found in~\cite{Reasoners}.

\paragraph*{Knowledge base satisfiability}

The most fundamental of the DL reasoning tasks is knowledge base
satisfiability.  The goal of this task is to decide whether or not a given
KB is satisfiable, that is, whether there is an interpretation that
satisfies all of its axioms.  Some important reasoning tasks, such as axiom
entailment, which we discuss next, can be reduced to KB satisfiability.

There are basically two kinds of approaches for checking KB satisfiability.
The \emph{model-theoretic} approaches try to construct a model for the KB or
to show that such construction must necessarily fail.  The
\emph{proof-theoretic approaches} manipulate the KB syntactically until a
contradiction is derived or until a point is reached where it is guaranteed
that no contradiction can be derived.  Examples of methods of the first kind
are the (tree) automata-based methods and the method of the tableaux.  As
examples of methods of the second kind we can cite the consequence-driven
methods and those based on resolution.

\paragraph*{Axiom entailment}

The goal of axiom entailment is to decide whether a given axiom $α$ is a
logical consequence of a given KB.  This problem can be reduced to the KB
satisfiability problem in DLs which support negation (such as \SROIQ).  In
DLs without such support this reduction is still possible but requires the
emulation of negation through some notion of opposition between axioms.

\paragraph*{Concept satisfiability and classification}

The goal of the concept satisfiability (consistency) is to decide whether a
concept $C$ is satisfiable with respect to a knowledge base $\KB$.  That is,
whether there is an interpretation $\I$ which satisfies $\KB$ and in which
$C^\I$ is populated (not empty).  Note that some concepts, such as $C⊓¬C$,
by definition cannot satisfied.

A related task is concept classification.  Its goal to decide whether the
concept names occurring in a KB can be put into a hierarchy according to
their subsumption ($⊑$) relationships.  This hierarchical ordering is often
a preliminary step executed before other reasoning tasks and also helps in
the visualization of KBs with a large number of GCIs.

\paragraph*{Named instance retrieval}

The named instance retrieval task takes a knowledge base $\KB$ and a concept
$C$ and returns all individual names $\T{a}$ such that $\KB⊨C(\T{a})$.  That
is, it determines all names of individuals which are instances of the
concept $C$ in every model of $\KB$.  This task can be extended to roles in
which case we search for pairs of individual names $⟨\T{a},\T{b}⟩$ such that
$\KB⊨r(\T{a},\T{b})$.

\paragraph*{Query answering}

The goal of query answering is to find an answer to a query $q$ in a given
knowledge base $\KB$.  The query $q$ is specified as a partial statement
whose missing parts (or blanks---actually, free variables) must filled with
individual names.  An answer to $q$ is a particular filling which makes
$\KB$ entail $q$, that is, a sequence of individual names which when
replaced for the blanks in $q$ makes $\KB⊨q$.  A natural extension of the
problem of finding an answer to a query is the problem of finding all
possible answers to a query.

\paragraph*{Induction and abduction}

The goal of induction is to analyze the assertional axioms (those in the
ABox) and generalize them by generating corresponding terminological axioms
(GCIs and RIAs).  The goal of abduction, on the other hand, is to determine,
given a knowledge base $\KB$ and an axiom $α$ not entailed by $\KB$, which
axioms (satisfying some basic requirements) need to be added into $\KB$ so
that $\KB⊨α$.  Differently from the previous tasks, both induction and
abduction are non-truth preserving tasks: the axioms they generate may be
falsified.

\subsection{Rules}
\label{sub:DL:rules}

Sometimes the reasoning tasks discussed above are not powerful enough.  Some
applications might need to represent relations and allow derivations which
go beyond what can be expressed and achieved with pure DL.  For instance,
relations and derivations involving arbitrary computations, such as those
involving arithmetic, or the derivation of facts exactly when some other
facts cannot be derived (non-monotonic inferences).

A common approach to improve the expressivity of a DL is to combine the
axioms in the KB with \emph{rules} expressed in some rule language.  A
popular choice of rule language are Horn clauses---a fragment of first-order
logic which is also the basis of the Prolog and Datalog programming
languages.  For example, the Semantic Web Rule Language (SWRL), pronounced
``swirl'', is essentially a combination of Datalog and the ontology language
OWL.  Other combinations of DL with rules which improve the expressivity of
the language (but in this case preserve decidability) are the Description
Logic Programs (DLP)~\cite{Grosof-B-N-2003} and Description Logic Rules
(DLR)~\cite{Krotzsch-M-2008}.  For more details on rules, we refer the
reader to~\cite{Hitzler-P-2010}.

\subsection{Other DLs}
\label{sub:DL:other-dls}

The DL family follows a convention which allows us to determine the features
of a specific DL from its name.  This convention is given by the scheme
\[
  ((\C{ALC}∣\C{S})\,[\C{H}]∣\C{SR})\,[\C{O}]\,[\C{I}]\,[\C{F}∣\C{N}∣\C{Q}],
\]
where parentheses denote grouping, the vertical bar stands for choice, and
the square brackets delimit optional components.  The meaning of the letters
in the above scheme is summarized in Table~\ref{tab:DL-naming} (adapted
from~\cite{Sikos-L-F-2017}).

\begin{table}[h]
  \caption{Features corresponding to the letters in a DL name.}
  \label{tab:DL-naming}
  \begingroup
  \small
  \begin{tabularx}{\textwidth}{lXp{16em}}
    \toprule
    Symbol & Available/missing features & Example\\
    \midrule
    $\C{ALC}$
    & Missing: RBox axioms, universal role, role inverses, cardinality
    constraints, nominal concepts, and self concepts.
    &$¬(\T{LiveAction}⊔\T{Animation})$\\\addlinespace
    $\C{S}$
    &$\C{ALC}$ plus role chain axioms of the form $\T{r}∘\T{r}⊑\T{r}$ for
    $\T{r}$ in $𝖭_R$.
    &$\T{partOf}∘\T{partOf}⊑\T{partOf}$\\\addlinespace
    $\C{H}$
    &$\C{ALC}$ plus role chain axioms of the form $r⊑s$.
    &$\T{remakeOf}⊑\T{basedOn}$\\\addlinespace
    $\C{SR}$
    &$\C{ALC}$ plus self concepts and all kinds of RBox axioms.
    &$\T{hasParent}∘\T{hasBrother}⊑\T{hasUncle}$\\\addlinespace
    $\C{O}$
    &Available: nominal concepts.
    &$\{\TT{universal},\T{disney}\}⊑\T{Studio}$\\\addlinespace
    $\C{I}$
    &Available: role inverses.
    &$\T{childOf}⊑\T{parentOf}^-$\\\addlinespace
    $\C{F}$
    &Available: role functionality statements (expressible as $⊤⊑{≤}1r.⊤$).
    &$⊤⊑{≤}1\T{officialWebsite}.⊤$\\\addlinespace
    $\C{N}$
    &Available: cardinality constraints of the form
    ${≥}nr.⊤$ and ${≤}nr.⊤$.
    &$\T{Series}⊑{≥}2.\T{hasEpisode}.⊤$\\\addlinespace
    $\C{Q}$
    &Available: all kinds of cardinality constraints.
    &$\T{Series}⊑{≥}2.\T{hasEpisode}.\T{Episode}$\\
    \bottomrule
  \end{tabularx}
  \endgroup
\end{table}

Besides \SROIQ\ other popular description logics are the logics $\C{ALC}$
and $\C{SHOIQ}$.  $\C{ALC}$~\cite{SchmidtSchauss-M-1991} is the smallest DL
which is boolean-closed, that is, in which boolean operators can be applied
to concepts without restrictions.  $\C{SHOIQ}$~\cite{Baader-F-2007} is
closely related to the ontology language OWL DL (version~1).  Note that both
$\C{ALC}$ and $\C{SHOIQ}$ are sublanguages of \SROIQ.

\paragraph*{Spatial and temporal extensions of DL}

In multimedia representation, we often need to describe spatial and temporal
relations between objects.  For instance, when annotating a particular scene
of a film we may need to specify the relative position of the characters in
the scene or the temporal sequence of their actions and dialogs.

\begin{figure}
  \centering
  \fbox{\includegraphics[width=.8\textwidth]{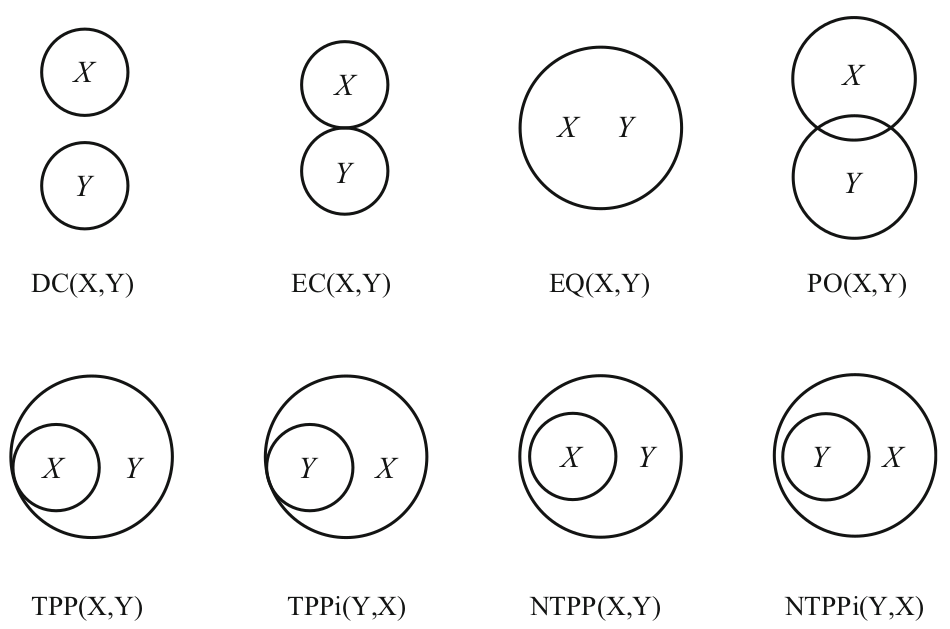}}
  \caption{The RCC8 relations~\cite{Sikos-L-F-2017}.}
  \label{fig:RCC8}
\end{figure}

A simple but expressive formalism for the specification of 2D spatial
relations is the \emph{region connection calculus} (RCC8).  The
RCC8~\cite{Randell-D-A-1992} describes eight basic relations which can take
place between two regions: disconnected (DC), externally connected (EC),
equal (EQ), partially overlapping (PO), tangential proper part (TPP),
tangential proper part inverse (TPPi), nontangential proper part (NTPP), and
nontangential proper part inverse (NTPPi).  These eight relations are
depicted in Figure~\ref{fig:RCC8}.  The
$\C{ALCI}_\text{RCC}$~\cite{Wessel-M-2003} is a family of spatial
description logics specifically designed to implement the RCC8 relations.
Other DLs which can be used for spatial reasoning are the logics
$\C{ALC}(\mathrm{F})$~\cite{Hudelot-C-2015} and
$\C{ALC}(\mathrm{CDC})$~\cite{Cristiani-M-2011}.  These two are more recent
proposals which are not based on the RCC.

To describe facts in time we can use one of the temporal DLs.  Most of these
are DL extensions with support for the usual operators of temporal logic,
such as the operators~$\lozenge$ (possibly in the future) and~$\square$
(always in the future).  The description logic
$\C{ALC}\text{-}\mathrm{LTL}$~\cite{Baader-F-2014} combines $\C{ALC}$ and
Linear Temporal Logic (LTL), which adopts a linear (non-branching) notion of
time and is usually employed in the specification of dynamic systems.
Another temporal DL is DL-CTL~\cite{Wang-Y-2014}; this combines DL with
Computational Tree Logic (CTL) which supports branching time.  There is also
an extension of the description logic $\C{SHIN}$ that when implemented in
OWL is called tOWL~\cite{Milea-V-2012}.  tOWL supports the representation of
time points and relations between time points and intervals.

\paragraph*{Fuzzy extensions of DL}

Information extracted from multimedia data is often ambiguous and imprecise.
But traditional DLs, which are fragments of classic first-order logic, have
no room for ambiguity or imprecision---the logic is boolean: either some
fact is true or it is false.  This means that when we use traditional DLs to
represent multimedia data we may lose some important information about
uncertainty.  Or worse, we may create some misleading sense of certainty
where no such certainty really exists.  One way to tackle this problem is to
represent uncertainty in the logic.

The fuzzy logics are particular kind of many-valued (non-boolean) logics
whose semantics involve a notion of \emph{truth degrees}.  These degrees
function as additional shades of gray between the boolean true and false,
which can be used to formalize vagueness and imprecision.  There are many
fuzzy extensions of DL.  We will not describe the here but we refer the
reader to~\cite{Borgwardt-S-2017}.




\section{Semantic Web Technologies}
\label{sec:semantic-web}

The term \emph{Semantic Web} refers to an effort initiated in the early
2000s to extend the traditional Web, then a global library of interlinked
HTML documents, to enable the semantic description and processing of these
documents.  Two decades later, the Web has become a much more diverse and
complex environment, and the success and feasibility of the Semantic Web
vision is somewhat debatable.  What is not debatable, though, is the success
of the standards and technologies developed in this period.  Nowadays, it
has become common to associate the term Semantic Web with these standards
and technologies.

\begin{figure}
  \centering
  \begin{tikzpicture}[
    box/.style={font=\small,draw,minimum height=1.8em},node distance=.3em]
    \node[box,minimum width=5em](OWL){\textbf{OWL}};
    \node[box,minimum width=5em,above=of OWL](SWRL){SWRL};
    \node[box,left=of OWL.west,minimum width=4.7em](SPARQL){\textbf{SPARQL}};
    \node[box,below=of OWL.south west,minimum width=10em](RDF)
    {\textbf{RDF}, RDFS};
    \node[box,below=of RDF,minimum width=10em](XML)
    {XML, Turtle, \dots};
    \node[box,below=of XML,minimum width=10em](URI){URI, IRI, Unicode};
  \end{tikzpicture}
  \caption{The Semantic Web stack.}
  \label{fig:stack}
\end{figure}
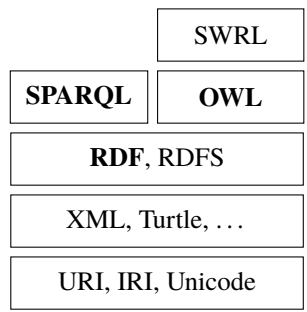

In this section, we will describe the core technologies of the Semantic Web.
These can be organized as a stack, as depicted in Figure~\ref{fig:stack},
where the technologies appearing on the top layers build on those which
appear on the bottom layers.  Although we will touch upon most of these
technologies, we will focus on the ones depicted in bold font in the figure,
namely, RDF, SPARQL, and OWL, with a special emphasis on the latter.

\subsection{RDF}
\label{sub:semantic-web:rdf}

One of the principal standards of the Semantic Web is the \emph{Resource
  Description Framework} (RDF)~\cite{W3C-RDF}.  RDF is a graph-based
representation language.  It provides a vocabulary and constructors for
describing \emph{sets of triples} denoting directed graphs.  These triples
are called \emph{s-p-o triples} and have the form
\[
  ⟨\text{\emph{subject}},\text{\emph{predicate}},\text{\emph{object}}⟩.
\]

Here are three s-p-o triples describing, respectively, the type, size in
bytes, and title associated with an audio file:
\[
  ⟨\text{song.mp3},\text{type},\text{Music}⟩,\
  ⟨\text{song.mp3},\text{size},10240⟩,\
  ⟨\text{song.mp3},\text{title},\text{``9th Symphony''}⟩.
\]
And here is the graph corresponding to these three s-p-o triples:
\begin{center}
  \begin{tikzpicture}[
    font=\small,node distance=2em,inner sep=1pt,outer sep=0,
    n/.style={draw,shape=ellipse,minimum width=6em}]
    \node[n](song){\strut song.mp3};
    \node[n,below=of song](size){\strut10240};
    \node[n,right=of size](title){\strut ``9th Symphony''};
    \node[n,left=of size](type){\strut Music};
    \begin{scope}[font=\footnotesize]
      \draw[->](song) -- node[auto]{size}(size);
      \draw[->](song) -- node[auto]{title}(title);
      \draw[->](song) -- node[above left]{type}(type);
    \end{scope}
  \end{tikzpicture}
\end{center}
Each triple stands for a directed edge in the graph.  The subject of the
triple becomes the source of the edge, the object becomes its target, and
the predicate becomes the edge label.

Of course, this is just an abstract example.  In an actual RDF graph the
components of the triples, that is, the labels of nodes and edges in the
graph, must obey a specific format (or \emph{syntax}).  A popular textual
syntax for RDF is Turtle (Terse RDF Triple Language)~\cite{W3C-Turtle}.
Turtle is an alternative to the widely supported but cumbersome RDF/XML
syntax.  All of our RDF examples will be written in Turtle.

There are three kinds of nodes in an RDF graph: IRI nodes, literal nodes,
and blank nodes.  An \emph{IRI node} is a node labeled by an IRI
(Internationalized Resource Identifier); an IRI is a generalization of URI
to support Unicode characters.  Here is an example IRI:
\[
  \text{\code|https://pt.wiktionary.org/wiki/maçã|}
\]
Note the non-ASCII characters ``\code|ç|'' and ``\code|ã|'' above.  In an
RDF graph, IRIs are used to denote arbitrary \emph{resources}, such as
physical objects, documents, images, abstract concepts, numbers, strings,
etc.

The second kind of node which can appear in an RDF graph is the literal
node.  A \emph{literal node} is any node labeled by a literal, that is, a
quoted string denoting a value in-place.  In the Turtle syntax, literals are
written enclosed in double quotes optionally accompanied with language tags
and type tags.  Here are three RDF literals (written in Turtle):
\begingroup
\vskip\abovedisplayskip
  \noindent\hfil
  \code|"Rock"|,\quad
  \code|"maçã"@pt|,\quad
  \code|"10240"^^xsd:integer|.
\vskip\belowdisplayskip
\endgroup
\noindent The first is a plain literal, the second is a literal with a
language tag (Portuguese in this case), and the third is a literal with a
type tag, also called typed literal.  The prefix \code|xsd:|, which occurs
in the third literal above, stands for the namespace where the type tag
\code|integer| is defined.  We will have more to say about namespaces in a
moment.

The third kind of node that can occur in an RDF graph is the blank node.  A
\emph{blank node} has no specific denotation; it is an unlabeled node which
is neither an IRI nor literal node.  Blank nodes are often used as dummy
nodes whose only purpose is to ease the encoding of complex structures in
graphs.

In an RDF triple $⟨s,p,o⟩$ the subject $s$ must be an IRI or blank node, the
predicate $p$ must be an IRI, and the object $o$ must be an IRI, literal, or
blank node.  When we assert an RDF triple we are saying informally that
``the relationship indicated by the predicate holds between the resource
denoted by the subject and the resource or value denoted by the object''.

Figure~\ref{fig:turtle} presents a possible Turtle encoding for the graph
describing the audio file ``song.mp3'' depicted a few paragraphs
ago.\footnotemark\ Each edge in the graph (that is, triple in the set)
becomes a corresponding triple in the Turtle document (lines~5--7).  The
subject, predicate, and object parts are separated by spaces and each triple
is terminated with a dot~(\code{.}).  Note that IRIs are written between
angle brackets \code[escapeinside=@@]{<@\dots@>}.

\footnotetext{Sometimes literals are represented as rectangles when
  depicting an RDF graph.  Incidentally, we did not follow this convention
  when depicting the graph describing ``song.mp3''.}

\begin{figure}[H]
\begin{minted}[bgcolor=bg,linenos=true]{turtle}
@base <http://example.org/data/> .
@prefix ex: <http://example.org/#> .
@prefix xsd: <http://www.w3.org/2001/XMLSchema#> .
<song.mp3> ex:type ex:Music .
<song.mp3> ex:size "10240"^^xsd:integer .
<song.mp3> ex:title "9th Symphony" .
\end{minted}
  \vskip-.5\baselineskip
\caption{Turtle encoding of the RDF triples describing ``song.mp3''.}
\label{fig:turtle}
\end{figure}

The directives in lines 1--3 of Figure~\ref{fig:turtle} define the
namespaces of the document.  That is, they specify how relative and prefixed
IRIs are to be interpreted.
The directive \code{@base} defines the base IRI of the whole Turtle
document.  Any relative IRI occurring in the document is to be interpreted
as relative to this base IRI.  So, for instance, the occurrences of
\code{<song.mp3>} in lines~4--6 of Figure~\ref{fig:turtle} are to be
interpreted as the absolute IRI:
\[
  \text{\code{http://example.org/data/song.mp3}}
\]

The directive \code{@prefix} introduces prefix labels which can be used to
abbreviate IRIs.  So, in Figure~\ref{fig:turtle}, the labels \code{ex:} and
\code{xsd:} stand for the following IRI prefixes:
\begin{minted}{text}
http://example.org/#
http://www.w3.org/2001/XMLSchema#
\end{minted}
Note that prefixed IRIs are \emph{not} written between angle brackets.

Turtle has a number of other features.  Some of these are useful for
avoiding tedious repetitions within the document.  For instance,
Figure~\ref{fig:turtle-2} uses some of these features to define in fewer
lines the same graph as Figure~\ref{fig:turtle}.  Note the semicolons
``\code{;}'' in lines~3 and~4 of Figure~\ref{fig:turtle-2}.  These are used
to separate predicate-object pairs referring to the same subject; in this case,
the IRI \code|http://example.org/data/song.mp3|.  This type of construction
is called a \emph{predicate list}.  Note also the typed literal 10240 in
line~4.  Some common types of literals, such as integers and floating-point
numbers, can be written in Turtle in their usual syntaxes.

\begin{figure}[H]
\begin{minted}[bgcolor=bg,linenos=true]{turtle}
@prefix ex: <http://example.org/#> .
<http://example.org/data/song.mp3>
  ex:type ex:Music ;
  ex:size 10240 ;
  ex:title "9th Symphony" .
\end{minted}
  \vskip-.5\baselineskip
\caption{Example of Figure~\ref{fig:turtle} rewritten using predicate lists.}
\label{fig:turtle-2}
\end{figure}

It is important to realize that an RDF graph is just a data structure, and
that Turtle, RDF/XML, and the other syntaxes for RDF are just ways to encode
this data structure in a string.  More importantly, the meanings of the
labels which occur in an RDF graph are not defined in the RDF standard.
These labels might indicate something to the person reading them, but to a
computer that understands only RDF these labels are meaningless
strings---they serve only to identify and distinguish specific components of
the graph.

There is another standard, called \emph{RDF Schema} (RDFS)~\cite{W3C-RDFS},
defined on top of RDF, which specifies a basic vocabulary (set predefined
terms) and associated meanings.  This vocabulary can be found in the
namespace:
\vskip\abovedisplayskip
  \hfil\code|http://www.w3.org/2000/01/rdf-schema#|
\vskip\belowdisplayskip
\noindent It provides a set of basic terms which can be used to define more
specialized terms whose meaning is given by describing their interrelations
with other terms.  (Note that we used a similar approach in
Section~\ref{sec:DL} when we introduced strings representing concepts and
roles, such as $\T{Director}$ and $\T{directs}$, and then defined their
meaning via GCIs and RIAs relating them with other terms.)

RDFS is sometimes called a \emph{lightweight ontology language} because it
allows us to specify ontologies (machine-processable specifications of
conceptualizations with a formally defined meaning) using a relatively small
vocabulary with a simple semantics.  This might be sufficient for certain
applications but, in general, sophisticated applications require more
powerful ontology languages.  The natural choice in this case is OWL, also a
Semantic Web technology.  We will describe OWL in
Section~\ref{sub:semantic-web:owl}.  Before that, however, let us present
SPARQL, a language for querying RDF graphs.

\subsection{SPARQL}
\label{sub:semantic-web:sparql}

In the previous section, we showed an RDF graph consisting of four nodes and
three edges.  When working with such small graphs one hardly needs to worry
about efficiency.  But, in practice, RDF graphs are rarely small.  It is not
uncommon to find applications that deal with graphs containing millions of
nodes and edges.  For such applications simple tools will not do.  They need
specialized tools, in particular, specialized databases, called \emph{graph
  databases} and \emph{triple stores}, designed to efficiently keep and
process large amounts of RDF data.

\emph{SPARQL} (SPARQL Protocol and RDF Query Language) is to graph databases
and triple stores what SQL is to relational databases.  It is a declarative
interface for querying and manipulating the contents of the
database/graph~\cite{W3C-SPARQL}.  There are two main kinds of SPARQL
queries: graph pattern matching and graph navigation.  We will discuss each
of these next.

\paragraph*{Graph pattern matching}

A graph pattern matching query is a query that searches for a given
(bounded) structural pattern in the graph.  In SPARQL, these structural
patterns are expressed as \emph{triple patterns}, which are RDF triples in
which the subject, predicate, or object may be a variable (name starting
with ``\code{?}'').  A \emph{basic graph pattern} (BGP) is the combination
of one or more triple patterns.

\begin{figure}
  \centering
  \begin{tikzpicture}[
    node distance=3em,
    n/.style={draw,shape=ellipse},
    l/.style={draw},
    font=\ttfamily\scriptsize
    ]
    \node[n](taxi-driver){:taxi-driver};
    \node[n,left=of taxi-driver,xshift=-2em](scorsese){:scorsese};
    \node[n,right=of taxi-driver,xshift=2em](de-niro){:de-niro};
    \node[n,above=of taxi-driver,xshift=-2em,anchor=east](Movie){:Movie};
    \node[l,above=of taxi-driver,xshift=2em,anchor=west](Title){``Taxi Driver''};
    \node[n,below=of taxi-driver](Person){:Person};
    \draw[->,bend left=10](scorsese) to node[above]{:actsIn} (taxi-driver);
    \draw[->,bend right=10](scorsese) to node[below]{:directs} (taxi-driver);
    \draw[->,bend right=25](scorsese) to node[below left]{:type} (Person);
    \draw[->](taxi-driver) to node[right]{:type} (Movie);
    \draw[->](taxi-driver) to node[right]{:title} (Title);
    \draw[->](de-niro) to node[above]{:actsIn} (taxi-driver);
    \draw[->,bend left=25](de-niro) to node[auto]{:type} (Person);
  \end{tikzpicture}
  \caption{RDF graph describing \emph{Taxi Driver}.}
  \label{fig:taxi-driver}
\end{figure}
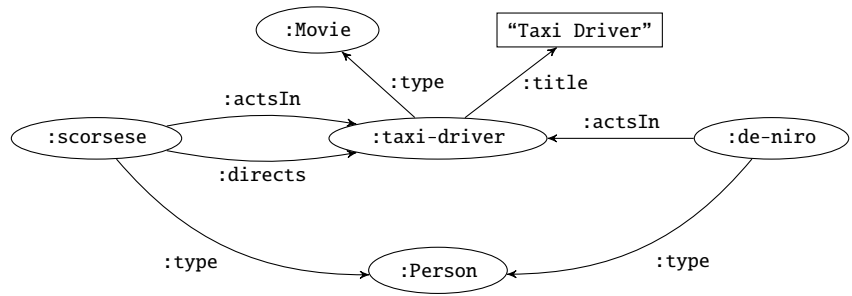

Consider the RDF graph depicted in Figure~\ref{fig:taxi-driver}, which
describes the film \emph{Taxi Driver}.  Suppose we want to list all pairs of
distinct persons who acted in this film.  We can specify this query in
SPARQL as follows:
\begin{minted}[linenos=true]{text}
PREFIX: <http://example.org/#>
SELECT ?x1 ?x2
WHERE {
  ?x1 :actsIn ?x3 . ?x1 :type :Person .
  ?x2 :actsIn ?x3 . ?x1 :type :Person .
  ?x3 :title "Taxi Driver" . ?x3 :type :Movie .
  FILTER (?x1 != ?x2)
}
\end{minted}
This query expresses a BGP consisting of six triple patterns.  It select all
pairs of labels $x_1$ and $x_2$ (line~2) with $x_1≠x_2$ (line~7) such
that, for some label $x_3$, each of the following triples occur in the
graph:
\begin{align*}
⟨x_1,\text{\code{:actsIn}},x_3⟩
&\quad\text{and}\quad
⟨x_1,\text{\code{:type}},\text{\code{:Person}}⟩\tag{line~4}\\
⟨x_2,\text{\code{:actsIn}},x_3⟩
&\quad\text{and}\quad
⟨x_2,\text{\code{:type}},\text{\code{:Person}}⟩\tag{line~5}\\
⟨x_3,\text{\code{:title}},\text{\code{"Taxi Driver"}}⟩
&\quad\text{and}\quad
⟨x_3,\text{\code{:type}},\text{\code{:Movie}}⟩\tag{line~6}
\end{align*}
If we apply this query to the graph of Figure~\ref{fig:taxi-driver} we get
the following result:
\begin{minted}[linenos]{text}
?x1         ?x2
---------------------
:scorsese   :de-niro
:de-niro    :scorsese
\end{minted}
That is, our \emph{answer set} consists exactly of two pairs of labels: one
where $x_1$ is $\text{\code{:scorsese}}$ and $x_2$ is
$\text{\code{:de-niro}}$ (line~3 in the result), and another where $x_1$ is
$\text{\code{:de-niro}}$ and $x_2$ is $\text{\code{:scorsese}}$ (line~4 in
the result).

\paragraph*{Graph navigation}

The other common kind of query supported by SPARQL are graph navigation
queries.  These are queries that navigate the topology of the graph through
paths of potentially arbitrary lengths.  A \emph{path query} is a basic
navigation query of the form $x\xrightarrow{α}y$ that retrieves all pairs
$⟨x,y⟩$ such that there is a path in the graph from $x$ to $y$ which
satisfies a given condition $α$.

For instance, the path queries
\[
  x\xrightarrow{\T{:directs}}y
  \quad\text{and}\quad
  x\xrightarrow{\T{:actsIn}\,∘\,\T{:type}}y
\]
search, respectively, for all nodes $x$ and $y$ such that there is an edge
labeled \code{:directs} from $x$ to $y$, and for all nodes $x$ and $y$ such
that there is path of size two from $x$ to $y$ where the first edge in the
path is labeled \code{:actsIn} and the second edge is labeled \code{:type}.
(Recall that a \emph{path} in a graph is a sequence of edges joining a
sequence of nodes.).

We can specify the latter query in SPARQL as follows:
\begin{minted}{text}
SELECT ?x ?y
WHERE  ?x (:actsIn/:type) ?y
\end{minted}
Here the symbol ``\code|/|'' stands for chaining and corresponds to the
``$∘$'' we used in the abstract notation.  If we apply this SPARQL query to
the graph of Figure~\ref{fig:taxi-driver} we get answer:
\begin{minted}{text}
?x          ?y
------------------
:scorsese  :Movie
:de-niro   :Movie
\end{minted}

For a more interesting example, consider the path query
\[
  x\xrightarrow{(\T{:actsIn}\,∘\,\T{:actsIn}^-)^+}y.
\]
This query searches for all actors who have a finite collaboration
distance\footnotemark, that is, actors $x$ and $y$ such that $x$ and $y$ are
co-stars in the same movie (distance 0), or $x$ and $y$ are co-stars of the same
$z$ in some movie (distance 1), or $x$ is co-star of some $z_1$ in some
movie and $y$ is co-star of some $z_2$ in some movie and $z_1$ and $z_2$ are
co-stars in the same movie (distance 2), and so on.  The following SPARQL code
implements this query:
\begin{minted}{text}
SELECT ?x ?y
WHERE  ?x (:actsIn/^:actsIn)+ ?y
\end{minted}
Here the symbol ``\code|^|'' denotes the inverse of an edge label (the
symbol ``$-$'' we used in the abstract notation) and ``\code|+|'' stands for
one or more repetitions of the pattern immediately preceding it.

\footnotetext{When $x$ is the actor Kevin Bacon this distance is known as
  the \emph{Bacon distance} or \emph{Bacon degree}.  See:
  \url{https://en.wikipedia.org/wiki/Six_Degrees_of_Kevin_Bacon}}

Suppose we add the following triples to the graph of
Figure~\ref{fig:taxi-driver}:
\[
  ⟨\text{\code{:de-niro}},\text{\code{:actsIn}},\text{\code{:cassino}}⟩
  \quad\text{and}\quad
  ⟨\text{\code{:sheron-stone}},\text{\code{:actsIn}},\text{\code{:cassino}}⟩.
\]
If we run the above SPARQL query over this updated graph we get the answer:
\begin{minted}{text}
?x              ?y
-----------------------------
:scorsese       :de-niro
:de-niro        :scorsese
:de-niro        :sharon-stone
:sharon-stone   :de-niro
:scorsese       :sharon-stone
:sharon-stone   :scorsese
\end{minted}

SPARQL has many more features, such as groups, options, alternatives,
filters, etc., which can be combined with basic graph patterns and
navigation operators in order to build complex queries.  Our goal here was
to provide a glimpse of SPARQL queries.  For a detailed introduction to
SPARQL, see~\cite{Hitzler-P-2010}.  Besides SPARQL, there are many other
languages for querying graphs.  See~\cite{Angles-R-2012} for a general
discussion of these.  For a practical introduction to graph databases and
triple stores, we recommend~\cite{Robinson-I-2013a}.

\subsection{OWL}
\label{sub:semantic-web:owl}

\emph{Web Ontology Language} (OWL) is a family of ontology languages whose
syntax and (formal) semantics is standardized~\cite{W3C-OWL2Overview}.
Recall that an \emph{ontology} is a set of precise descriptive statements
about some part of the world---statements like those of description logic.
OWL is based on description logic but also has many features which go beyond
pure DL, such as datatypes and features for versioning and annotating
ontologies.  Here we will focus on version~2 of OWL (called OWL~2).  This is
the most recent version and it is fully compatible with the previous
version.

There are three sublanguages of OWL, also called \emph{species}: OWL Full,
OWL DL, and OWL Lite.  OWL Full contains both OWL DL and OWL Lite, and OWL
DL contains OWL Lite.  What distinguishes these three sublanguages is their
degree of expressivity and, consequently, the computational cost of their
associated reasoning tasks.

As with any language, in order to present OWL we will need to pick a
particular syntax.  Here we will use the RDF/Turtle syntax.  Any OWL
document written in this syntax is also a valid RDF document.  Besides RDF,
other common syntaxes for OWL are the functional-style syntax and the
Manchester syntax.  However, the only syntax that is mandatory to be
supported by all OWL~2 tools is the RDF/XML syntax~\cite{W3C-OWLPrimer}.

Next, we will present the specific terms and semantics of OWL via
translations from DL.  (An OWL~2 DL ontology is essentially an \SROIQ\
knowledge base.)  Before doing that, however, we must clear up some
terminological conflicts.  For historical reasons, OWL and DL sometimes use
different terms to refer to the same objects.  These terminological
differences are summarized in Table~\ref{tab:terms}.

\begin{table}
  \centering
  \caption{Terminological differences between OWL and DL.}
  \label{tab:terms}
  \begin{tabular}{lll}
    \toprule
    OWL                  & DL\\
    \midrule
    class name           & concept name\\
    class                & concept\\
    object property name & role name\\
    object property      & role\\
    ontology             & knowledge base\\
    axiom                & axiom\\
    vocabulary           & vocabulary/signature\\
    \bottomrule
  \end{tabular}
\end{table}

\paragraph*{From \SROIQ\ to OWL~2 DL}

The translation of an \SROIQ\ knowledge base into an OWL document consists
of three steps.\footnotemark\ First, we need to prefix to the document the
following namespace declarations:
\begin{minted}{turtle}
@prefix owl: <http://www.w3.org/2002/07/owl#> .
@prefix rdfs: <http://www.w3.org/2000/01/rdf-schema#> .
@prefix rdf: <http://www.w3.org/1999/02/22-rdf-syntax-ns#> .
@prefix xsd: <http://www.w3.org/2001/XMLSchema#> .
\end{minted}
The prefixes \code|rdfs:|, \code|rdf:|, and \code|xsd:| are necessary because
OWL reuses terms from these namespaces.

\footnotetext{The translations we present here were adapted from Sebastian
  Rudolph's excellent text~\cite{Rudolph-S-2011}.}

The second thing we need to do is add to the OWL document typing statements
for the concept names and role names that occur in the KB.  That is, for
each concept name $\T{C}$ occurring in the KB, we add to the document the
triple:
\begin{minted}[escapeinside=@@]{turtle}
@$\T{C}$@ rdf:type owl:Class .
\end{minted}
And for each role name $\T{r}$ occurring in the KB, we add the triple:
\begin{minted}[escapeinside=@@]{turtle}
@$\T{r}$@ rdf:type owl:ObjectProperty .
\end{minted}
Figure~\ref{fig:movie-facts-owl} shows the OWL document corresponding to the
\SROIQ\ movie facts KB presented at the end of Section~\ref{sub:DL:syntax}.
The typing statements in lines~7--13 were generated by the above
translations.  Note that due to the \code|@prefix| directive in line~5, IRIs
with an empty prefix label such as \code|:Actor| are to be interpreted as
prefixed by the IRI \code|http://example.org/movie-facts#|.

The third step to translate an \SROIQ\ knowledge base to OWL is the most
complicated of the three: we need to translate the axioms that occur in each
of the boxes.  The complexity in this case stems from the rich syntactical
structure of GCIs and RIAs, which cannot be directly encoded in RDF.  To
rewrite GCIs and RIAs in RDF we will need to use blank nodes and techniques
for encoding lists in graphs.

The precise translation of arbitrary \SROIQ\ axioms into OWL is given by
function~$⟦⬚⟧$ below:
\begingroup
\interdisplaylinepenalty=100
\begin{align*}
  ⟦r_1∘⋯∘r_n⊑r⟧
  &=⟦r⟧_{\ssR}\ \TSC{owl:propertyChainAxiom}\
    \TC{(}\ ⟦r_1⟧_{\ssR}…⟦r_n⟧_{\ssR}\ \TC{)}\ \TC{.}\\
  ⟦\SS{Sym}(r)⟧
  &=⟦r⟧_{\ssR}\ \TSC{owl:type owl:SymmetricProperty}\ \TC{.}\\
  ⟦\SS{Asy}(r)⟧
  &=⟦r⟧_{\ssR}\ \TSC{owl:type owl:AsymmetricProperty}\ \TC{.}\\
  ⟦\SS{Tra}(r)⟧
  &=⟦r⟧_{\ssR}\ \TSC{owl:type owl:TransitiveProperty}\ \TC{.}\\
  ⟦\SS{Ref}(r)⟧
  &=⟦r⟧_{\ssR}\ \TSC{owl:type owl:ReflexiveProperty}\ \TC{.}\\
  ⟦\SS{Irr}(r)⟧
  &=⟦r⟧_{\ssR}\ \TSC{owl:type owl:IrreflexiveProperty}\ \TC{.}\\
  ⟦\SS{Dis}(r,s)⟧
  &=⟦r⟧_{\ssR}\ \TSC{owl:propertyDisjointWith}\ ⟦s⟧_{\ssR}\ \TC{.}\\
  ⟦C⊑D⟧
  &=⟦C⟧_{\ssC}\ \TSC{rdfs:subClassOf}\ ⟦D⟧_{\ssC}\ \TC{.}\\
  ⟦C(\T{a})⟧
  &=\T{a}\ \TSC{rdf:type}\ ⟦C⟧_{\ssC}\ \TC{.}\\
  ⟦\T{r}(\T{a},\T{b}⟧
  &=\T{a}\ \T{r}\ \T{b}\ \TC{.}\\
  ⟦\T{r}^-(\T{a},\T{b})⟧
  &=\T{b}\ \T{r}\ \T{a}\ \TC{.}\\
  ⟦¬r(\T{a},\T{b})⟧
  &=\TC{[]}\
    \begin{aligned}[t]
      &\TC{rdf:type owl:NegativePropertyAssertion}\ \TC{;}\\
      &\TC{owl:assertionProperty}\ ⟦r⟧_{\ssR}\ \TC{;}\\
      &\TC{owl:sourceIndividual}\ \T{a}\ \TC{;}\\
      &\TC{owl:targetIndividual}\ \T{b}\ \TC{.}
    \end{aligned}\\
  ⟦\T{a}≈\T{b}⟧
  &=\T{a}\ \TSC{owl:sameAs}\ \T{b}\ \TC{.}\\
  ⟦\T{a}≉\T{b}⟧
  &=\T{a}\ \TSC{owl:differentFrom}\ \T{b}\ \TC{.}
\end{align*}
\endgroup
where the auxiliary functions $⟦⬚⟧_{\ssR}$ and $⟦⬚⟧_{\ssC}$ are used to
translate \SROIQ\ roles and concept expression.  These auxiliary functions
are defined as follows:
\begingroup
\interdisplaylinepenalty=100
\begin{align*}
  ⟦u⟧_{\ssR}
  &=\TSC{owl:topObjectProperty}\\
  ⟦\T{r}⟧_{\ssR}
  &=\T{r}\\
  ⟦\T{r}^-⟧_{\ssR}
  &=\TC{[}\ \TSC{owl:inverseOf :}\T{r}\ \TC{]}\\[4\jot]
  ⟦\T{C}⟧_{\ssC}
  &=\T{C}\\
  ⟦⊤⟧_{\ssC}
  &=\TSC{owl:Thing}\\
  ⟦⊥⟧_{\ssC}
  &=\TSC{owl:Nothing}\\
  ⟦\{\T{a}_1,…,\T{a}_n\}⟧_{\ssC}
  &=\TC{[}\ \TSC{rdf:type owl:Class ; owl:oneOf }
    \TC{(}\ \TSC{:}\T{a}_1\ …\ \TSC{:}\T{a}_n\ \TC{)]}\\
  ⟦¬C⟧_{\ssC}
  &=\TC{[}\ \TSC{rdf:type owl:Class ; owl:complementnOf}\ ⟦C⟧_{\ssC}\ \TC{]}\\
  ⟦C_1⊓⋯⊓C_n⟧_{\ssC}
  &=\TC{[}\ \TSC{rdf:type owl:Class ; owl:intersectionOf}\
    \TC{(}\ ⟦C_1⟧_{\ssC}\ …\ ⟦C_n⟧_{\ssC}\ \TC{)}\TC{]}\\
  ⟦C_1⊔⋯⊔C_n⟧_{\ssC}
  &=\TC{[}\ \TSC{rdf:type owl:Class ; owl:unionOf}\
    \TC{(}\ ⟦C_1⟧_{\ssC}\ …\ ⟦C_n⟧_{\ssC}\ \TC{)}\TC{]}\\
  ⟦∃{r}.C⟧_{\ssC}
  &=\TC{[}\
    \begin{aligned}[t]
      &\TSC{rdf:type owl:Restriction ; }\\
      &\TSC{owl:onProperty}\ ⟦r⟧_{\ssR}\
      \TSC{; owl:someValuesFrom}\ ⟦C⟧_{\ssC}\ \TC{]}
    \end{aligned}\\
  ⟦∀{r}.C⟧_{\ssC}
  &=\TC{[}\
    \begin{aligned}[t]
      &\TSC{rdf:type owl:Restriction ; }\\
      &\TSC{owl:onProperty}\ ⟦r⟧_{\ssR}\
      \TSC{; owl:allValuesFrom}\ ⟦C⟧_{\ssC}\ \TC{]}
    \end{aligned}\\
  ⟦∃{r}.\S{Self}⟧_{\ssC}
  &=\TC{[}\
    \begin{aligned}[t]
      &\TSC{rdf:type owl:Restriction ; }\\
      &\TSC{owl:onProperty}\ ⟦r⟧_{\ssR}\ \TSC{; owl:hasSelf true}\ \TC{]}
    \end{aligned}\\
  ⟦{≥}nr.C⟧_{\ssC}
  &=\TC{[}\
    \begin{aligned}[t]
      &\TSC{rdf:type owl:Restriction ; }\\
      &\TSC{owl:minQualifiedCardinality}\ n\ \TSC{;}\\
      &\TSC{owl:onProperty}\ ⟦r⟧_{\ssR}\ \TSC{; owl:onClass}\ ⟦C⟧_{\ssC}\ \TC{]}
    \end{aligned}\\
  ⟦{≤}nr.C⟧_{\ssC}
  &=\TC{[}\
  \begin{aligned}[t]
      &\TSC{rdf:type owl:Restriction ; }\\
      &\TSC{owl:maxQualifiedCardinality}\ n\ \TSC{;}\\
      &\TSC{owl:onProperty}\ ⟦r⟧_{\ssR}\ \TSC{; owl:onClass}\ ⟦C⟧_{\ssC}\ \TC{]}
    \end{aligned}
\end{align*}
\endgroup

Two remarks on the above translation are in order.  First, note that
individual names, concept names, and role names are kept as is by $⟦⬚⟧$,
$⟦⬚⟧_{\ssR}$, and $⟦⬚⟧_{\ssC}$.  For simplicity, we assume that these names
are defined in the namespace of the empty prefix.

Second, note that complex role and concept expressions are translated by
$⟦⬚⟧_{\ssR}$ and $⟦⬚⟧_{\ssC}$ as subgraphs identified by a blank node.  In
Turtle, these blank nodes are introduced by the square brackets.  So, for
instance, the role~$\T{r}^-$ becomes the subgraph
\begin{center}
  \begin{tikzpicture}[
    n/.style={draw,shape=circle},
    node distance=8em,
    font=\ttfamily\scriptsize]
    \node[n,minimum width=2em,minimum height=2em](b){};
    \node[n,right=of b,minimum width=2em,minimum height=2em](r){$\T{r}$};
    \draw[->](b) to node[above]{owl:inverseOf} (r);
  \end{tikzpicture}
\end{center}
in which the blank node on the left-hand side stands for the inverse of $\T{r}$.
Similarly, the complex concept ${≥}2\T{knows}.\T{Actor}$ is translated as
the subgraph
\begin{center}
  \begin{tikzpicture}[
    n/.style={draw,shape=ellipse,minimum height=2em},
    node distance=4em,
    font=\ttfamily\scriptsize]
    \node[n](restriction){owl:Restriction};
    \node[n,shape=rectangle,right=of restriction](2){2};
    \node[n,right=of 2](knows){knows};
    \node[n,right=of knows](Actor){Actor};
    \node[n,minimum width=2em,shape=circle,yshift=6em](b)
    at ($(restriction.west)!.5!(Actor.east)$){};
    \draw[->,bend right=20](b) to node[above left]{rdf:type}(restriction);
    \draw[->,bend right=15](b) to node[left,text width=5em,align=right]
    {owl:min\\Qualified\\Cardinality}(2);
    \draw[->,bend left=15](b) to node[right,xshift=.75em,text width=4em,
    align=left]
    {owl:on\\Property}(knows);
    \draw[->,bend left=20](b) to node[above right]{owl:onClass}(Actor);
  \end{tikzpicture}
\end{center}
Again, the blank node above stands for the complex concept as a whole.

The result of using function $⟦⬚⟧$ to translate the axioms in the movie
facts KB is depicted in Figure~\ref{fig:movie-facts-owl}.  The translated
ABox axioms are listed in lines 16--28, the TBox axioms in lines~31--53, and
the RBox axioms in lines 55--56.

\begin{figure}[p]
  \centering
\begin{minted}{turtle}
@prefix owl:    <http://www.w3.org/2002/07/owl#> .
@prefix rdfs:   <http://www.w3.org/2000/01/rdf-schema#> .
@prefix rdf:    <http://www.w3.org/1999/02/22-rdf-syntax-ns#> .
@prefix xsd:    <http://www.w3.org/2001/XMLSchema#> .
@prefix :       <http://example.org/movie-facts#> .

:Actor      rdf:type    owl:Class .
:Director   rdf:type    owl:Class .
:Person     rdf:type    owl:Class .
:actsIn     rdf:type    owl:ObjectProperty .
:directs    rdf:type    owl:ObjectProperty .
:knows      rdf:type    owl:ObjectProperty .
:likes      rdf:type    owl:ObjectProperty .

# Abox
:kubrick    rdf:type    :Director .
:de-niro    rdf:type    [rdf:type owl:Class ;
                         owl:intersectionOf (:Actor :Director)] .
:stanley    rdf:type    [rdf:type owl:Restriction ;
                         owl:minQualifiedCardinality 2 ;
                         owl:onProperty :knows ; owl:onClass :Actor ] .
:kubrick    :directs    :space-odyssey .
[]          rdf:type    owl:NegativePropertyAssertion ;
                        owl:assertionProperty :directs ;
                        owl:sourceIndividual :kubrick ;
                        owl:targetIndividual :taxi-driver .
:kubrick    owl:sameAs  :stanely .
:kubrick    owl:differentFrom   :de-niro .

# TBox
:Director   rdfs:subClassOf     [rdf:type owl:Restriction ;
                                 owl:onProperty :directs ;
                                 owl:someValuesFrom owl:Thing] .
[rdf:type owl:Restriction ;
 owl:onProperty :directs ;
 owl:someValuesFrom owl:Thing]
            rdfs:subClassOf     :Director .
:Director   rdfs:subClassOf     :Person .
:Director   rdfs:subClassOf     [rdf:type owl:Restriction ;
                                 owl:minQualifiedCardinality 1 ;
                                 owl:onProperty :knows ;
                                 owl:onClass :Actor] .
[rdf:type owl:Class ;
 owl:intersectionOf  ([rdf:type owl:Restriction ;
                       owl:onProperty :directs ;
                       owl:someValuesFrom owl:Thing]
                       [rdf:type owl:Restriction ;
                        owl:onProperty :directs ;
                        owl:allValuesFrom [rdf:type owl:Class ;
                                           owl:oneOf (:a-bronx-tale
                                                      :good-shepherd)]])]
            rdfs:subClassOf     [rdf:type owl:Class ;
                                 owl:oneOf (:de-niro)] .
# RBox
:likes  owl:propertyChainAxiom  (:directs) .
:knows  owl:propertyChainAxiom  (:directs [owl:inverseOf :actsIn]) .
\end{minted}
  \vskip-.5\baselineskip
  \caption{OWL corresponding to the movie facts KB of Section~\ref{sub:DL:syntax}.}
  \label{fig:movie-facts-owl}
\end{figure}

\paragraph*{Emulating the additional axiom types of OWL in \SROIQ}

The features of OWL used by the previous translation algorithm are
sufficient for representing any \SROIQ\ axiom in OWL.  But OWL has many more
features, including further types of logical axioms which have no direct
counterpart in \SROIQ.  The extra types of logical axioms of OWL are useful
in practice, but they do not add new capabilities from the point of view of
the logic.  That is, they are just a syntactic sugar that can always be
rewritten in terms of the types of axioms available in \SROIQ.
See~\cite{Rudolph-S-2011} for details.

\paragraph*{Tools for manipulating ontologies}

If we load the Turtle file of Figure~\ref{fig:movie-facts-owl} on
Protégé\footnotemark\ and ask for a graph of the ontology, we get the
following:
\begin{center}
  \includegraphics[width=.8\textwidth]{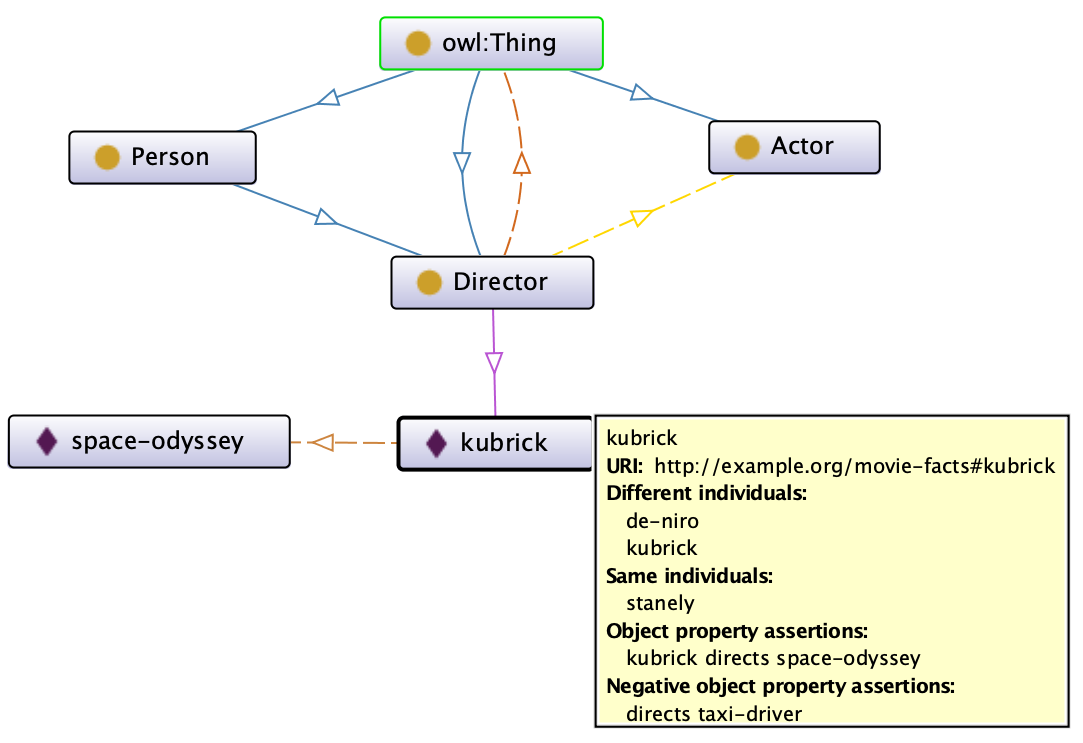}
\end{center}
\vskip-.5\baselineskip
\footnotetext{\url{https://protege.stanford.edu/}}
Here the nodes with a yellow circle stand for the concepts (classes), the
nodes with purple diamonds stand for individuals, and the colored arrows
denote specific relationships between them.  For instance, the blue arrow
from \code{Person} to \code{Director} represents the ``has subclass''
relationship.  The yellow box above, with additional information about
\code{kubrick}, is the tool-tip displayed by Protégé when we hover the mouse
cursor over a particular node or edge of the graph.

Protégé is the most widely used ontology editor.  It supports many plugins
for ontology visualization, debugging, and reasoning.  Besides Protégé,
there are many other tools for designing and implementing ontologies.  For a
comprehensive list, see~\cite{Hitzler-P-2010} and~\cite{Tools}.

\paragraph*{Linked data and knowledge graphs}

Over the years the Semantic Web vision has changed its focus from the
description of linked documents to the description of linked data in
general.  The term \emph{linked open data} (LOD) has been used to refer to
\emph{open} datasets of linked data, such as
DBpedia\footnote{\url{https://wiki.dbpedia.org/}} (derived from Wikipedia),
Wikidata\footnote{\url{https://www.wikidata.org/}} (curated by the Wikimedia
Foundation), etc.  Most of these datasets adopt the technologies we have
been discussing, but they are usually not ontologies in the strict sense.

Another term often associated with linked data is \emph{knowledge graph}.
This is a general term which alludes to the fact that the system or database
to which the term is applied uses a graph to represent knowledge.  In this
sense, an OWL document encoded in RDF is a knowledge graph.  Being a
knowledge graph does not imply being an ontology.  For instance, the graph
depicted in Figure~\ref{fig:taxi-driver} is a knowledge graph, but it is
certainly not an ontology.

\paragraph*{Ontologies for multimedia}

There are numerous ontologies for describing multimedia.  We can classify
these broadly into three groups: descriptor-derived, general purpose, and
specialized ontologies.  The ontologies in the first group are those derived
from classic metadata descriptors standards, such as MPEG-7, MPEG-21, and
TV-Anytime.  These ontologies are complex, limited in terms of expressivity,
and have design issues (they often do not follow good practices of ontology
engineering).
The ontologies on the second group tend to be more mature and well-defined.
On the other hand, as those in the first group, the ontologies in the third
group are also problematic.  They often have conceptualization problems, for
instance, their is scope is unclear, and usually do not reuse terms from
more general (upper-level) ontologies, which is considered bad practice.

Table~\ref{tab:mm-ontologies} lists some of the ontologies in the three
groups.  For more details on these, such as references, a discussion of
their pros and cons, etc., see~\cite{Sikos-L-F-2017}.

\begin{table}
  \small
  \centering
  \begin{tabular}{lll}
    \toprule
                    &Vocabulary/Ontology                &Language\\
    \midrule
    Descriptor-derived
                    &Content Ontology for the TV-Anytime Content &OWL\\
                    &Content Ontology for the TV-Anytime Format  &OWL\\
                    &Core Ontology for Multimedia (COMM)         &OWL\\
                    &Visual Descriptor Ontology (VDO)            &OWL\\
    \midrule
    General purpose &Dublin Core                        &RDFS\\
                    &FOAF                               &OWL\\
                    &Schema.org                         &OWL\\
                    &SUMO                               &SUO-KIF\\
                    &WordNet 3.x                        &OWL\\
    \midrule
    Specialized     &3D Modeling Language (3DMO)        &OWL~2\\
                    &Audio Effects Ontology (AUFX-O)    &OWL\\
                    &Linked Movie Database (LMD)        &RDFS\\
                    &Multimedia Metadata Ontology (M3O) &OWL\\
                    &Ontology for Media Resources       &OWL\\
                    &Video Ontology (VidOnt)            &OWL\\
    \bottomrule
  \end{tabular}
  \caption{Some vocabularies and ontologies for multimedia.}
  \label{tab:mm-ontologies}
\end{table}

The point of using one of these open ontologies is, of course, to reuse
their terms and definitions instead constructing everything from scratch.
For instance, using Schema.org and Dublin Core we can rewrite the RDF
describing the audio file ``song.mp3'' (Figure~\ref{fig:turtle-2}) as
follows:
\begin{minted}{text}
@prefix dc: <http://purl.org/dc/elements/1.1/> .
@prefix rdf: <http://www.w3.org/1999/02/22-rdf-syntax-ns#> .
@prefix schema: <http://schema.org/> .
<http://example.org/data/song.mp3> rdf:type schema:AudioObject ;
                                   schema:contentSize "10240" ;
                                   dc:title "9th Symphony" .
\end{minted}

Note that we can use URI fragments~\cite{Pfeiffer-S-2012} to refer to
specific parts of the resource being described.  For instance, if we want to
indicate that the key of some particular segment, say from 10s to 20s, of
``song.mp3'' is C Major, we can write:
\begin{minted}{text}
@prefix keys: <http://purl.org/NET/c4dm/keys.owl#>
@prefix mo:   <http://purl.org/ontology/mo/>
<http://example.org/data/song.mp3#t=10,20> mo:key keys:CMajor .
\end{minted}
where \code{mo:key} and \code{keys:CMajor} are terms from the Music
Ontology\footnote{http://musicontology.com/}.  Similarly, we can refer to
spatial fragments of resources using URI fragments of the form
\code[escapeinside=@@]|#xywh=@\emph{x},\emph{y},\emph{w},\emph{h}@|.  We
will have more to say about the description of media objects and their parts
in next section.




\section{A Hybrid Approach: Hyperknowledge}
\label{sec:HK}


IBM Hyperlinked Knowledge Graph~\cite{Moreno-M-2017}, or Hyperknowledge for
short, is a hybrid model for knowledge representation that permits the
specification and interlinking of data objects and concepts.  It extends the
Nested Context Model (NCM)~\cite{Soares-L-F-G-2005}, a classic hypermedia
model, with constructs that combine features from the domains of
\emph{hyper}media and \emph{knowledge} representation.  To better understand
why the Hyperknowledge model was proposed, let us briefly discuss the main
concerns of these two communities.

The hypermedia community has been concerned with the definition of models
and languages for expressing relationships among media objects (texts,
images, videos, etc.).  In languages such as HTML, NCL, SMIL, and SVG we can
specify how different objects interact with each other and with users but we
cannot link these objects to the concepts they represent---at least not
directly.  Also, we cannot describe in these languages ontologies like those
discussed in Section~\ref{sec:DL}.  Metadata standards (such as MPEG-7,
MPEG-21, SMPTE, EXIF, etc.) have been proposed as a solution to combine the
media content and semantics, but these standards usually focus on low-level
features, such as codecs, bit-rates, color spaces, etc.

The knowledge representation community, on the other hand, has focused on
the development of models for representing facts about the world and on
methods for querying and inferring things from these models.  Applications
in the knowledge engineering domain tend to assume that knowledge bases are
composed purely of facts and often disregard any (multimedia) data that
might be related to these facts.  For instance, we can use RDF or OWL for
encoding facts in a knowledge base, but the multimedia content described by
these facts is usually kept in separated bases, which are handled by
different tools and systems.

The goal of the Hyperknowledge model is to unify hypermedia and knowledge
representation, and with this to enable the development of both
semantic-aware multimedia applications and multimedia-aware knowledge bases.

\subsection{Hyperknowledge in a nutshell}

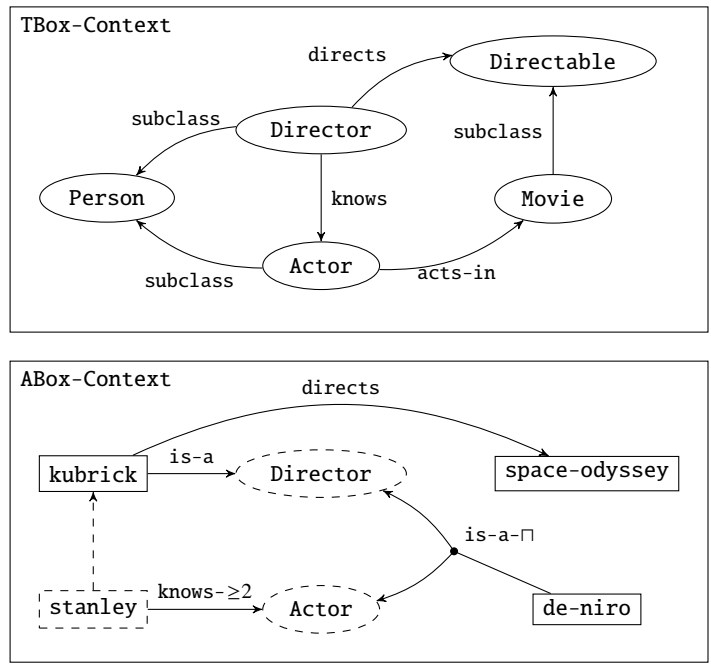
\begin{figure}
  \centering
  \begin{tikzpicture}[font=\ttfamily\footnotesize,
    n/.style={draw,ellipse},
    b/.style={draw},
    l/.style={font=\ttfamily\scriptsize}]
    \begin{scope}[node distance=3em,local bounding box=box]
      \node[n](Directable){Directable};
      \node[n,below=of Directable](Movie){Movie};
      \node[n,left=of {$(Directable)!.5!(Movie)$},xshift=-2em]
      (Director){Director};
      \node[n,below=of Director](Actor){Actor};
      \node[n,left=of {$(Director)!.5!(Actor)$},xshift=-2em](Person){Person};
      \draw[->,bend right=20](Director)
      to node[l,above,yshift=2pt,xshift=-2pt]{subclass} (Person);
      \draw[->,bend left=20](Actor)
      to node[l,below,yshift=-2pt,xshift=-2pt]{subclass} (Person);
      \draw[->](Director) to node[l,auto]{knows} (Actor);
      \draw[->,bend left=20](Director) to node[l,auto]{directs} (Directable);
      \draw[->,bend right=20](Actor) to node[l,below]{acts-in} (Movie);
      \draw[->](Movie) to node[l,auto]{subclass} (Directable);
      \node[n,dashed,anchor=north](DirectorA)
      at ($(Director)+(0,-11)$) {Director};
      \node[n,dashed,below=of DirectorA](ActorA){Actor};
      \node[b,left=of DirectorA](kubrick){kubrick};
      \node[b,dashed](stanley) at (kubrick|-ActorA){stanley};
      \node[b,right=of DirectorA](space-odyssey){space-odyssey};
      \node[b](de-niro) at (space-odyssey|-ActorA){de-niro};
      \draw[->](stanley) to node[l,auto]{knows-${≥}2$} (ActorA);
      \draw[->](kubrick) to node[l,auto]{is-a} (DirectorA);
      \draw[->,dashed](stanley) to (kubrick);
      \draw[->,bend left=25](kubrick) to node[l,above](K){directs} (space-odyssey);
      \coordinate(X) at ($(ActorA)!.5!(de-niro)$);
      \coordinate(A) at ($(X)+(0,2)$);
      \fill(A) circle (1.5pt);
      \node[l,above right] at (A){is-a-$⊓$};
      \draw (de-niro) to (A);
      \draw[->,bend right=15](A) to (DirectorA);
      \draw[->,bend left=15](A) to (ActorA);
    \end{scope}
    \coordinate(W) at ($(Actor.south)!.5!(K.north)$);
    \draw ($(box.north west)+(-1,1)$) node[anchor=north west]{TBox-Context}
    rectangle ($(box.south east|-W)+(1,0)$);
    \draw ($(box.north west|-W)+(-1,-1)$) node[anchor=north west]{ABox-Context}
    rectangle ($(box.south east)+(1,-1)$);
  \end{tikzpicture}
  \caption {A Hyperknowledge base corresponding to the movie facts KB.}
  \label{fig:hk-kb}
\end{figure}

The entities of the Hyperknowledge model are inherited from the
NCM~\cite{Soares-L-F-G-2005}.  These entities comprise anchors, (atomic)
nodes, compositions, links, and connectors.  We will introduce each of these
entities through an example.  Figure \ref{fig:hk-kb} shows a possible
representation in Hyperknowledge of a part of the movie facts KB, presented
at the end of Section~\ref{sec:DL}.  For simplicity, here we adopt the
closed world assumption, that is, we assume that facts not represented in
the figure are false.  (We discussed this assumption and the more general
open world assumption in Section~\ref{sub:DL:reasoning}.)

A Hyperknowledge \emph{context} is a container element that groups related
elements and relationships.  The characterization of related entities and
the decision of grouping them into contexts is a modeling decision which
depends on the application.  In Figure~\ref{fig:hk-kb}, the contexts are
depicted as the bounding boxes enclosing the other elements.  There are two
contexts in this figure.  Context \code{TBox-Context} contains the elements
and relationships that correspond to the TBox of the movie facts KB.  The
other context, \code{ABox-Context}, contains the elements and relationships
corresponding to the ABox.

The ellipses and smaller rectangles in Figure~\ref{fig:hk-kb} stand for
atomic nodes.  These can represent either concepts (ellipses) or data
(rectangles).  Here we use the word concept in the same sense in which it is
used in DL.  That is, a \emph{concept} (or class in OWL terminology)
represents an abstract characteristic of certain individuals.  In
Figure~\ref{fig:hk-kb}, all nodes in \code{TBox-Context}, namely,
\code{Person}, \code{Director}, \code{Actor}, \code{Movie}, and
\code{Directable}, are concept nodes.

The other kind of atomic nodes which occur in Figure~\ref{fig:hk-kb} are
data nodes.  A \emph{data node} is a node containing some data content.  In
the figure, the data nodes are the rectangles labeled \code{de-niro},
\code{kubrick}, and \code{space-odyssey}, which occur in the
\code{ABox-Context}.  The idea in this example is that data nodes stand for
multimedia content.  For instance, \code{de-niro} and \code{kubrick} might
refer to a JPEG picture of the corresponding movie directors, and
\code{space-odyssey} might refer to an Ogg video of the corresponding film.


The dashed nodes in Figure~\ref{fig:hk-kb} are neither concept nodes nor
data nodes; they stand for \emph{references}.  That is, they are nodes that
refer to other nodes, allowing that the latter be reused in other situations
(contexts).  So, for instance, the concept nodes \code{Actor} and
\code{Director}, which are defined in the TBox context, are reused in the
ABox context.  Similarly, node \code{stanley} of the ABox context is a
reference to the node \code{kubrick} in the same context.  The idea here is
that anything which operates over \code{stanley} is actually operating over
\code{kubrick}.

This type of indirection is useful in cases where the same entity is
associated with different names in possibly different contexts.  For
instance, the singer and song-writer Bob Dylan has recorded some folk songs
under the pseudonym of Blind Boy Grunt.  Any references to Blind Boy Grunt
are essentially references to Bob Dylan.  But the distinction between them
might be important for some applications.

Back to Figure~\ref{fig:hk-kb}, the arrows in the figure stand for links
representing relationships between nodes.  Each link is associated with a
connector (not represented in the figure) which defines the link type.  A
connector defines an abstract relation and associates with it a label and
zero or more restrictions.  A link can be seen as an instance of its
connector, that is, as an actual realization of the abstract relation
defined by the connectors.

In the figure, the link in the TBox context between \code{Actor} and
\code{Person} labeled ``subclass'' expresses the fact that every instance of
\code{Actor} is also an instance of a \code{Person}.  In other words, it
encodes the TBox axiom $\T{Actor}⊑\T{Person}$.  Similarly, the link
``directs'' between \code{Director} and \code{Directable} encodes the TBox
axiom $\T{Director}⊑∃\T{directs}.⊤$.

Unlike the edges of RDF graphs, which are necessarily binary, Hyperknowledge
links can connect more than two entities.  For instance, in
Figure~\ref{fig:hk-kb} the link ``is-a-$⊓$'' in the ABox context connects
\code{de-niro} to both \code{Actor} and \code{Director}, and serves to
encode the ABox axiom $\T{Actor}⊓\T{Director}(\TT{de-niro})$.

\subsection{HSL and HyQL}

The Hyperknowledge model is accompanied by a specification language, called
HSL, and a declarative query language, called HyQL.  We will now present
both of these technologies.

\paragraph*{HSL}

HSL (Hyperknowledge Specification Language) is a language for the
specification of hyperknowledge representations, which can be stored in
hyperknowledge bases, or HK bases.  Figure~\ref{fig:movie-facts-hsl} depicts
the HSL document in JSON syntax corresponding to the HK base of
Figure~\ref{fig:hk-kb}.

\begin{figure}[p]
  \centering
\begin{minted}{text}
["hsl", [
    ["connector", "subclass", {"type": "fact"}, [
        ["role", "subject"],
        ["role", "object"]]],
    ["connector", "is-a", {"type": "fact"}, [
        ["role", "subject"],
        ["role", "object"]]],
    ["connector", "is-a-cap", {"type": "fact"}, [
        ["role", "subject"],
        ["role", "object1"],
        ["role", "object2"]]],
    @$\ldots$@

    ["context", "TBox-Context", [
        ["node", "Actor", {"type": "Concept"}],
        ["node", "Directable", {"type": "Concept"}],
        ["node", "Director", {"type": "Concept"}],
        ["node", "Person", {"type": "Concept"}],
        ["node", "Movie", {"type": "Movie"}],

        ["link", {"connector": "subclass"}, [
            ["bind", {"subject": "Director"}],
            ["bind", {"object": "Person"}]]],

        ["link", {"connector": "subclass"}, [
            ["bind", {"subject": "Actor"}],
            ["bind", {"object": "Person"}]]]
        @$\ldots$@
    ]],

    ["context", "ABox-Context", null, [
        ["node", "kubrick", {"type": "Data",
                             "uri": "http://example.org/kubrick.jpg"}],
        ["node", "stanley", {"type": "Reference",
                             "refer": "kubrick"}],
        ["node", "Director", {"type": "Reference",
                              "refer": "TBox-Context/Director"}],
        ["node", "Actor", {"type": "Reference",
                           "refer": "TBox-Context/Actor"}],
        ["node", "space-odyssey", {"type": "Data",
                                   "uri": "http://example.org/odyssey.ogv"}],
        ["node", "de-niro", {"type": "Data",
                             "uri": "http://example.org/de-niro.jpg"}],

        ["link", {"connector": "is-a"}, [
            ["bind", {"subject": "kubrick"}],
            ["bind", {"object": "Director"}]]],

        ["link", {"connector": "is-a-cap"}, [
            ["bind", {"subject": "de-niro"}],
            ["bind", {"object1": "Director"}],
            ["bind", {"object2": "Actor"}]]]
        @$\ldots$@
    ]]
]]
\end{minted}
  \vskip-.5\baselineskip
  \caption{Partial HSL corresponding to the HK base of Figure~\ref{fig:hk-kb}.}
  \label{fig:movie-facts-hsl}
\end{figure}

An HSL document consists of a tree of elements.  In the JSON syntax, each
element is an array of the form:
\begin{minted}[escapeinside=@@]{text}
[@{\rm\emph{tag}}@, @{\rm\emph{id}}@, {@{\rm\emph{attrs}}@}, [@{\rm\emph{children}}@]]
\end{minted}
where \emph{tag} is the element tag (``hsl'', ``connector'', ``context'',
``node'', etc.), \emph{id} is the element id, \emph{attrs} is a key-value
map of element attributes, and \emph{children} is a list of child elements.
Depending on the \emph{tag}, any of the last three components, \emph{id},
\emph{attrs}, or \emph{children}, are optional.  The meaning and allowed
values for each of these components also varies from tag to tag.  Instead of
giving the full syntax of HSL, we will introduce its main elements using
Figure~\ref{fig:movie-facts-hsl} as a guide.

The root element of an HSL document is element ``hsl'' (line~1 of
Figure~\ref{fig:movie-facts-hsl}).  Three kinds of elements may appear
inside the root: connectors, contexts, and atomic nodes.  A connector
specifies the format of the links of the document.  For each distinct link
(arrow) label in Figure~\ref{fig:hk-kb} there is a connector in the HSL
document.  For instance, the connectors with ids ``subclass'', ``is-a'', and
``is-a-cap'' (lines~2--11) define the homonymous relations and are
referenced by the links on lines~21--25 and 45--49.

Each context (bounding box) in Figure~\ref{fig:hk-kb} becomes a ``context''
element in the HSL document (lines~14--29 and~31--54) and the concepts
(ellipses) and data objects (rectangles) of Figure~\ref{fig:hk-kb} become
``node'' elements inside the corresponding contexts.  For instance, the
concept \code{Director} in the TBox context becomes the node with id
``Director'' and type attribute ``Concept'' (line~17).  The data object
\code{kubrick} in the ABox context becomes the node with id ``kubrick'',
type ``Data'', and content URI ``http://example.org/kubrick.jpg''
(lines~32--33).  Note that node references---dashed ellipses and dashed
rectangles in Figure~\ref{fig:hk-kb}---become nodes of type ``Reference''
(lines~34--39) with the ``refer'' attribute defining the id of the target
node.

The links (arrows) of Figure~\ref{fig:hk-kb} become ``link'' elements in the
HSL document.  Each link must reference one of the connectors defined just
below the root element of the document.  For instance, the link (arrow) from
\code{Director} to \code{Person} with label \code{subclass} in
Figure~\ref{fig:hk-kb} becomes the link element defined on lines~21--23
whose attribute ``connector'' has value ``subclass'', which is the id of the
connector that defines the subclass relation (lines~2--4).  Each element
``role'' in the connector, in this case the roles ``subject'' and ``object''
(lines~3 and~4) of connector ``subclass'', must be matched by a
corresponding bind in the link, in this case the ``bind'' elements on
lines~22--23, which establish that in this particular instantiation of
connector ``subclass'' node ``Director'' is the subject and node ``Person''
is the object.  The omitted connectors and links in the HSL document of
Figure~\ref{fig:movie-facts-hsl} (lines~12, 28, and~53) all follow a similar
pattern.

\paragraph*{HyQL}

\newcommand*{\hN}[1]{\text{\emph{#1}}}
\newcommand*{\hT}[1]{\text{`{\ttfamily\textbf{#1}}'}}
\newcommand*{\hW}{\hN{name}}

From a graph-theoretic point of view, the Hyperknowledge model supports
advanced features, such as nested nodes (contexts) and hyperedges ($n$-ary
links), which are not present in traditional knowledge graphs.  A language
for querying Hyperknowledge graphs, thus, has to cope with these features
and with the peculiarities of the data which is stored in a Hyperknowledge
base.  For instance, queries involving multimedia objects often need to
refer to temporal or spatial fragments of these objects and to correlate
objects and fragments in time and space.

HyQL (Hyperknowledge Query Language) is a declarative language for querying
hyperknowledge graphs.  A HyQL query specifies a complex
Hyperknowledge-graph pattern, possibly involving path navigation, level
shifts, aggregation, filters, and specific features or fragments of the
underlying data.
The complete syntax of HyQL is presented in Figure~\ref{fig:hyql}.  We will
not discuss this syntax and its accompanying semantics in detail, though.
Instead, we will introduce HyQL gradually, through example queries over the
HK base of Figure~\ref{fig:hk-kb}.

\begin{figure}[p]
\footnotesize
\begin{framed}
\begin{align*}
  \hN{query}&\Coloneqq
  \hN{varlist}\ \hT{select}\ [\hN{modifier}]\ \hN{targetlist}\
  [\hT{where}\ \hN{clause}]\\
  \hN{varlist}&\Coloneqq
  \hN{vardecl}\ \{\hT{,}\ \hN{vardecl}\}\\
  \hN{vardecl}&\Coloneqq
  \hT{let}\ \hN{var}\\
  \hN{var}&\Coloneqq
  \hW\ \hT{as}\ \hW\mid\hW\ \hT{:}\ \hW\mid\hW\\
  \hN{modifier}&\Coloneqq
  \hT{distinct}\mid\hT{count}\mid\hT{max}\mid\hT{min}\mid\hT{sum}\mid\hT{avg}\\
  \hN{targetlist}&\Coloneqq
  \hN{target}\ \{\hT{,}\ \hN{target}\}\\
  \hN{target}&\Coloneqq
  \hN{node}\mid\hN{nodeattr}\mid\hN{nodeanchor}\mid\hN{funcall}\\
  \hN{node}&\Coloneqq
  \hN{var}\\
  \hN{nodeattr}&\Coloneqq
  \hN{var}\,\hT{.}\,\hN{var}\\
  \hN{nodeanchor}&\Coloneqq
  \hN{var}\,\hT{\#}\,\hN{var}\\
  \hN{funcall}&\Coloneqq
  \hW\ \hT{(}\,\hN{explist}\,\hT{)}\\
  \hN{explist}&\Coloneqq
  \hN{exp}\ \{\hT{,}\ \hN{exp}\}\\
  \hN{exp}&\Coloneqq
  \hN{boolean}\mid\hN{number}\mid\hN{string}\mid\hN{node}\mid\hN{nodeattr}\\
  \hN{clause}&\Coloneqq
  \hN{linkclause}\mid\hN{timeclause}\mid\hN{spaceclause}\mid\hN{contextclause}\\
  &\quad\mid\hN{spoclause}\mid\hN{attrclause}\mid\hN{funclause}\mid\hN{anchorclause}\\
  &\quad\mid\hN{clause}\ \hT{and}\ \hN{clause}\\
  \hN{linkclause}&\Coloneqq
  \hN{var}\ \hT{[}\ \{\hW\ \hT{:}\ \hN{var}\}\ \hT{]}\\
  \hN{timeclause}&\Coloneqq
  \hN{var}\ \hN{timeop}\ \hN{var}\\
  \hN{spaceclause}&\Coloneqq
  \hN{var}\ \hN{spaceop}\ \hN{var}\\
  \hN{contextclause}&\Coloneqq
  \hN{var}\ \hT{from}\ \hN{var}\\
  \hN{spoclause}&\Coloneqq
  \hN{var}\ \hN{var}\ \hN{var}\\
  \hN{attrclause}&\Coloneqq
  \hN{nodeattr}\ \hN{binop}\ \hN{exp}\\
  \hN{funclause}&\Coloneqq
  \hN{funcall}\ \hN{binop}\ \hN{exp}\\
  \hN{anchorclause}&\Coloneqq
  \hN{nodeanchor}\ \hN{timeop}\ \hN{nodeanchor}\\
  &\quad\mid\hN{nodeanchor}\ \hN{timeop}\ \hN{var}\\
  &\quad\mid\hN{nodeanchor}\ \hN{spaceop}\ \hN{nodeanchor}\\
  &\quad\mid\hN{nodeanchor}\ \hN{spaceop}\ \hN{var}\\
  &\quad\mid\hN{nodeanchor}\ \hN{var}\ \hN{var}\\
  \hN{timeop}&\Coloneqq
  \hT{before}\mid\hT{after}\mid\hT{meets}\mid\hT{overlaps}\mid\hT{start}\\
  &\quad\mid\hT{during}\mid\hT{finishes}\mid\hT{equals}\\
  \hN{spaceop}&\Coloneqq
  \hT{side}\mid\hT{above}\mid\hT{bellow}\mid\hT{collides}\mid\hT{contains}\\
  \hN{binop}&\Coloneqq
  \hT{<}\mid\hT{<=}\mid\hT{>}\mid\hT{>=}\mid\hT{==}\mid\hT{!=}
\end{align*}
\end{framed}
  \vskip-.5\baselineskip
  \caption{The complete syntax of HyQL.  $\{\hN{A}\}$ means zero or more
    $\hN{A}$s, and $[\hN{A}]$ means an optional $\hN{A}$.}
  \label{fig:hyql}
\end{figure}

Here is a HyQL query that lists all instances of concept ``Director'' in the
HK base of Figure~\ref{fig:hk-kb}:
\begin{minted}{text}
select Director
\end{minted}
The results in this case are ``kubrick'', ``stanely'', and ``de-niro''.  And
here is a query that counts the number of actors in the HK base:
\begin{minted}{text}
select count Actor
\end{minted}
The answer in this case is 1, as just ``de-niro'' is an instance of the
concept ``Actor''.

HyQL queries can also contain filters.  For instance, the query
\begin{minted}{text}
select Director
where Director directs space-odyssey
\end{minted}
contains a filter (\code{where}-clause) that filters out only the directors
who have directed (link ``directs'') the node ``space-odyssey''.  The answer
in this case is ``kubrick'' and ``stanley''.
Filters can be combined with other filters and also test node attributes.
For instance, the query
\begin{minted}{text}
select Director
where Director.uri == "http://example.org/kubrick.jpg"
      and Director.type != "Reference"
\end{minted}
selects all nodes which are instances of ``Director'' such that their
``uri'' attribute is equal to ``http://example.org/kubrick.jpg'' and their
type is not ``Reference''.  The answer in this case is only ``kubrick''.

We conclude the discussion of HyQL with a query that compares the spatial
position of objects in a picture.  For this query, we will need to introduce
the notion of anchor~\cite{s2019general}.  An anchor specifies a part of a
data node.  Anchors can be the source and target of links and can also be
independently designated via node references.

Consider the following HSL snippet:
\begin{minted}[linenos=true]{text}
["node", "picture", {"type": "Data", "uri": "marat.jpg"}, [
    ["anchor", "knife",
     {"x": "58px", "y": "766px", "w": "160px", "h": "37px"}],
    ["anchor", "marat",
     {"x": "39px", "y": "334px", "w": "146px", "h": "111px"}]]],
["node", "Knife", {"type": "Concept"}],
["node", "Person", {"type": "Concept"}],
["link", {"connector": "is-a"}, [
    ["bind", {"subject": "picture#knife"}],
    ["bind", {"object": "Knife"}]]],
["link", {"connector": "is-a"}, [
    ["bind", {"subject": "picture#marat"}],
    ["bind", {"object": "Person"}]]],
\end{minted}
This HSL snippet declares three nodes: ``picture'' (lines~1--5), which is a
data node whose URI points to the picture of Figure~\ref{fig:marat}, and
``Knife'' (line~6) and ``Person'' (line~7) which are concept nodes.  The
anchor elements (lines~2--3 and 4--5) of the ``picture'' node specify
rectangular regions in the picture, which correspond roughly to the regions
labeled ``Jean-Paul Marat'' and ``Knife'' in Figure~\ref{fig:marat}.  The
idea here is that we can use anchors to describe parts of nodes, in this
case regions of picture, and then use links to connect regions to their
semantic descriptions.  In the above snippet, the links connect the region
which depicts a knife with the concept ``Knife'' (lines~8--10) and the
region which depicts a person with the concept ``Person'' (lines~11--13).

Now suppose that our HK base contains many more pictures whose regions are
semantically described as above.  We can then use HyQL to find precisely
those images in which a certain spatial relationship between the concepts
represented in the image hold.  For instance, the HyQL query
\begin{minted}{text}
select Picture where Person above Knife
\end{minted}
lists all pictures in which a region described as a person appears above a
region described as a knife.  Clearly, the previous data node ``picture''
(line~1) matches this query.

HyQL has many more features, such as temporal operators (for comparing
temporal fragments of nodes), arithmetic operators, path and navigation
operators, etc.  The description of these operators is beyond the scope of
this chapter.  We refer the reader to~\cite{8665691,8527459} for additional
examples of HyQL queries involving spatial and temporal anchors.

We now shift the discussion the tools which allow us to use the
Hyperknowledge model in practice.

\subsection{Hyperknowledge tools}

There are two main tools for developing applications based on
Hyperknowledge, namely, the Hyperknowledge Base (HKBase) and the Knowledge
Explorer System (KES).

\paragraph*{HKBase}

\begin{figure}[h!]
  \centering
  \includegraphics[scale=0.45]{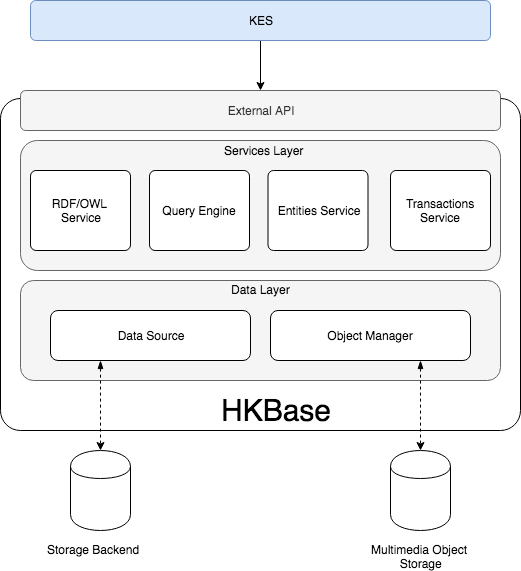}
  \caption{HKBase architecture.}
  \label{fig:hkbase}
\end{figure}

The \emph{HKBase} is a hybrid database system based on Hyperknowledge.  By
hybrid we mean that it stores multimedia objects besides statements about a
given domain.  Additionally, it has been designed from scratch to use
Hyperknowledge as both internal data model and external API.  For
compatibility reasons, the HKBase also has features for importing and
exporting RDF and OWL.  In the case of importation, these statements are
converted to Hyperknowledge entities before being stored in the underlying
database.  Figure~\ref{fig:hkbase} depicts the overall architecture of the
HKBase.

The HKBase architecture consists of three layers.  The first one is the
external API which exposes a REST API to applications.  The next layer is
the services layer which implements the core functionalities of the HKBase.
Within this layer, the RDF/OWL service handles the conversion among those
formats and Hyperknowledge.  The query engine is the component that
processes queries written in HyQL.  The entities service is responsible for
creating, retrieving, updating and deleting Hyperknowledge entities.  And
the transaction service manages requests that should be executed within a
transaction.

The third and last layer of the HKBase is the data layer which handles low
level access to storage back-ends.  Its data source component is responsible
for persisting in a given database the structured statements represented in
Hyperknowledge.  And its object manager component deals with the storage of
multimedia objects.

\paragraph*{KES}

\emph{KES} (Knowledge Explorer System)~\cite{kes} is an application built on
top of the HKBase.  Using KES, users can visualize the content stored in a
Hyperknowledge base as well as curate it by adding, updating, or deleting
entities and relationships.  The system is collaborative, that is, it
supports multiple users working simultaneously on the same repository
through different devices.  KES also has facilities for importing and
manipulating OWL and RDF files.

\begin{figure}
  \centering
  \includegraphics[width=\textwidth]{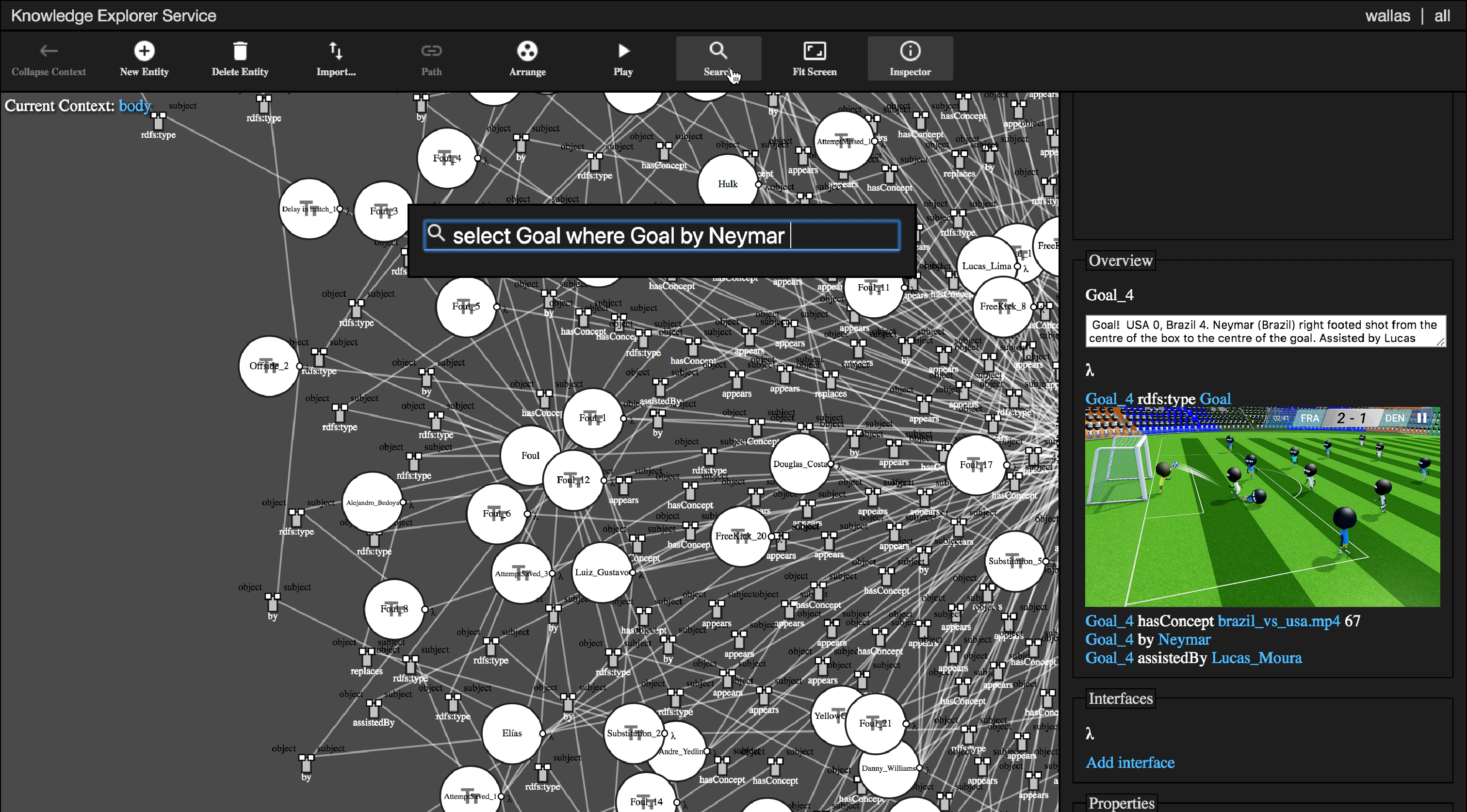}
  \caption{Screenshot of the user interface of KES.  (Soccer image by Abdul
    Rahman, CC BY-NC-ND 4.0.)}
  \label{fig:kes}
\end{figure}

Figure~\ref{fig:kes} shows a screenshot of KES.  The main canvas shows a
graph-based representation of the contents of a Hyperknowledge base.
Besides Hyperknowledge entities, KES permits the direct manipulation of
multimedia content.  For instance, the screenshot of Figure~\ref{fig:kes}
shows that the user has run the query: ``Select Goal where Goal by Neymar''.
(The Hyperknowledge base, in this case, contains facts about the soccer
domain~\cite{soccer}.)  This query returns all goals scored by the player
called ``Neymar''.  Note that the answer set display by KES contains all
Hyperknowledge entities that satisfy the query, that is, all media nodes
which are instances of \code{Goal} and which have at least one link stating
that the goal was scored by \code{Neymar}.  The content of these media nodes
is displayed on the right-hand side of the screen.

\section{Further Reading}
\label{sec:further-reading}

In this chapter, we presented an overview of symbolic AI and discussed its
application to multimedia.  Like any introduction, this one was necessarily
incomplete, especially given the breadth of the topics covered.  Here are
some suggestions for further reading which complement those given along the
text:

\paragraph*{Logic}

A classic introduction to mathematical logic is~\cite{Enderton-H-B-2001}.
The primary reference for description logic is~\cite{Baader-F-2007}.  A good
introduction to DL through \SROIQ\ is given in~\cite{Rudolph-S-2011}.  For
modal logics, especially temporal logic, a good starting-point
is~\cite{Goldblatt-R-1992}.  For a discussion of DL and multimedia,
see~\cite{Sikos-L-F-2017}.

\paragraph*{Semantic Web}

For a seminal discussion of the Semantic Web vision,
see~\cite{BernersLee-T-2001}.  An overview of the principles and
technologies of the Semantic Web is given in~\cite{Hitzler-P-2010}.
Specifically for OWL, a good place to start is~\cite{W3C-OWLPrimer}.  For a
comprehensive treatment of ontologies, see~\cite{Staab-S-2009}.

\paragraph*{Hyperknowledge}

The primary reference for Hyperknowledge is~\cite{Moreno-M-2017}.  Some
further references describing particular applications of Hyperknowledge to
document understanding and temporal reasoning are~\cite{Moreno-M-2018a} and
\cite{8527459}.




\bibliographystyle{acm}
\bibliography{main}
\end{document}